\pdfoutput=1
\documentclass[11pt]{article}

\usepackage[]{naacl2021}

\usepackage{times}
\usepackage{latexsym}

\usepackage[T1]{fontenc}
\usepackage[utf8]{inputenc}

\usepackage{microtype}

\usepackage{amsmath}
\usepackage{amssymb}
\usepackage{mathtools}
\usepackage{arydshln}
\usepackage{booktabs}
\usepackage{tabularx}
\usepackage{multirow}
\usepackage{tikz}
\usepackage{pgfplots}
\usepackage[normalem]{ulem}
\usepackage{xcolor}
\usepackage{dashbox}%
\usepackage{textcomp}
\usepackage{tcolorbox}
\usepackage{adjustbox}
\usepackage{lipsum}
\usepackage[inline]{enumitem}

\pgfplotsset{compat=1.13}

\definecolor{c0}{cmyk}{1,0.3968,0,0.2588} 
\definecolor{c1}{cmyk}{0,0.6175,0.8848,0.1490} 
\definecolor{c2}{cmyk}{0.1127,0.6690,0,0.4431} 
\definecolor{c3}{cmyk}{0.6765,0.2017,0,0.0667} 
\definecolor{c4}{cmyk}{0.3081,0,0.7209,0.3255} 
\definecolor{c5}{cmyk}{0,0.8765,0.7099,0.3647} 
\definecolor{cwhite}{cmyk}{0,0,0,0}
\definecolor{darkgrey}{RGB}{180,180,180}
\definecolor{decentgrey}{RGB}{220,220,220}
\usetikzlibrary{calc,fit,positioning,arrows,intersections}
\usepgfplotslibrary{fillbetween}

\pgfdeclarelayer{bg}
\pgfsetlayers{bg,main}

\tikzset{
	keep name/.style={
		prefix after command={
			\pgfextra{\let\fixname\tikzlastnode}
		}
	},
	partialbox/.style={
		keep name,
		append after command={
			(\fixname.north) -- 
			(\fixname.north west) -- 
			(\fixname.south west) -- 
			([xshift=-#1]\fixname.south)
			(\fixname.north) -- 
			(\fixname.north east) -- 
			(\fixname.south east) -- 
			([xshift=#1]\fixname.south)
		}
	},
	partialbox/.default=5pt
}

\newtcbox{\inlinepattern}{on line,colback=c0!10,colframe=white,size=fbox,arc=3pt, box align=base,before upper=\strut,
	top=-4pt, bottom=-4pt, boxrule=0pt}
\newtcbox{\pattern}{on line,colback=c0!10,colframe=white,size=fbox,arc=3pt, box align=base,before upper=\strut,
top=-2pt, bottom=-2pt, boxrule=0pt}
\newtcolorbox{multipattern}{on line,colback=c0!10,colframe=white,size=fbox,arc=3pt, box align=base, top=-2pt, bottom=0pt, boxrule=0pt, before=\adjustbox{valign=c}\bgroup, after=\egroup, before upper=\strut}

\DeclareMathOperator*{\argmax}{arg\,max}

\newcolumntype{Y}{>{\centering\arraybackslash}X}

\newcommand{\gpt}{GPT\nobreakdash-3}
\newcommand{\pet}{\textsc{Pet}}
\newcommand{\ipet}{i\textsc{Pet}}

\newcommand\mask{\_\_}
\newcommand{\et}{\fontshape{it}\selectfont}
\newcommand{\bt}{\fontseries{b}\selectfont}
\newcommand{\compellipsis}{\mathinner{{\ldotp}{\ldotp}{\ldotp}}}
\newcommand{\pzero}{\phantom{0}}
\newcommand{\negphantom}[1]{\settowidth{\dimen0}{#1}\hspace*{-\dimen0}}
\newcommand\sbullet[1][.5]{\mathbin{\vcenter{\hbox{\scalebox{#1}{$\bullet$}}}}}

\title{It's Not Just Size That Matters: \\ Small Language Models Are Also Few-Shot Learners}

\author{
	Timo Schick$^{1,2}$ \and Hinrich Sch\"utze$^{1}$ \\[0.5em]
	$^{1}$ Center for Information and Language Processing, LMU Munich, Germany \\
	$^{2}$ Sulzer GmbH, Munich, Germany \\[0.5em]
	{\tt timo.schick@sulzer.de}
}

\date{}

\newcounter{notecounter}

\newcommand{\enoteson}{\long\gdef\enote##1##2{{
\stepcounter{notecounter}
{\large\bf
\hspace{1cm}\arabic{notecounter} $<<<$ ##1: ##2
$>>>$\hspace{1cm}}}}}
\enoteson
\begin{document}
\maketitle
\begin{abstract}
When scaled to hundreds of billions of parameters,
pretrained language models such
as \gpt{} \citep{brown2020language} achieve remarkable
few-shot performance. 
                      However, enormous amounts of compute
                      are required for training and applying
                      such big models, resulting in
                      a large carbon footprint and making
                      it  difficult for researchers and
                      practitioners to use them. We show
                      that performance similar to \gpt{} can
                      be obtained with language models that
                      are much ``greener'' in that their
                      parameter count is several orders of
                      magnitude smaller. This is achieved by
                      converting textual inputs into cloze
                      questions that contain a task
                      description, combined with
                      gradient-based optimization;
                      exploiting unlabeled data gives
                      further improvements. We identify  key
                      factors required for successful
                      natural language understanding with
                      small language models.\footnote{Our implementation is publicly available at \url{https://github.com/timoschick/pet}.}
\end{abstract}

\section{Introduction}

Pretraining ever-larger language models (LMs) on massive   corpora has led to large
improvements in NLP
\cite[\emph{i.a.}]{radford2018improving,devlin2018bert,liu2019roberta,raffel2019exploring}. A standard approach is to replace the pretrained model's output layer with a task-specific head and finetune the entire model on a set of labeled training data. However, language modeling is not only a powerful pretraining objective, but many tasks can be reformulated as cloze questions (e.g., by appending phrases such as ``the correct answer is \mask{}''), allowing pretrained LMs to solve them without any or with only very few labeled examples \cite{radford2018language,schick2020exploiting}.

\begin{figure}
	\begin{tikzpicture}
	\begin{axis}[
	cycle list name=color list,
	xlabel={\sffamily\small Parameters (Millions)},
	ylabel={\sffamily\small SuperGLUE Performance},
	axis line style={decentgrey!95!black},
	grid=major,
	major grid style={line width=.2pt,draw=decentgrey},
	ymin = 45,
	ymax = 80,
	xmin = 100,
	xmax = 1000000,
	xmode = log,
	minor tick style={decentgrey!0},
	major tick style={decentgrey},
	log basis x={10},
	xtick pos=left,
	ytick pos=left,
	ylabel near ticks,
	xlabel near ticks,
	xticklabels={$10^2$, $10^3$, $10^4$, $10^5$, $10^6$},
	tick align=outside,
	tick label style={font=\footnotesize},
	major tick length=0.075cm,
	width = \linewidth,
	height = 0.23\textheight,
	log ticks with fixed point,
	x tick label style={/pgf/number format/1000 sep=\,},
	]
	\addplot[mark=*, c0, thick, mark options={solid}] coordinates {
		(125,50.1)
		(350,56.2)
		(760,56.8)
		(1300,60.0)
		(2700,64.3)
		(6700,63.6)
		(13000,66.9)
		(175000,73.2) 
	} node[right,pos=1,xshift=0.025cm]{\small\sffamily GPT-3};
	
	\addplot[mark=*, c1, thick, mark options={solid}] coordinates {
		(223,74.1)
	} node[right,pos=1,xshift=0.025cm]{\small\sffamily PET};
	
	\addplot[mark=*, c2, thick, mark options={solid}] coordinates {
		(223,76.8)
	} node[right,pos=1,xshift=0.025cm]{\small\sffamily iPET};
	
	\end{axis}
	\end{tikzpicture}
	\caption{Performance on SuperGLUE with 32 training examples.
\textbf{ALBERT with \pet/\ipet{}
outperforms \gpt{} although it is much ``greener'' in that it
          has three orders of magnitude
          fewer parameters.}}
	\label{figure:intro}
\end{figure}
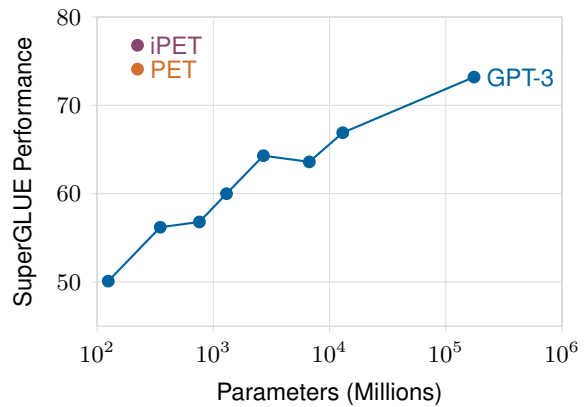 

Recently, \citet{brown2020language} introduced \gpt{}, a
pretrained LM with an enormous 175 billion parameters, and
showed that it has amazing few-shot abilities: By
reformulating tasks as LM problems, \gpt{} achieves near
state-of-the-art results for some
SuperGLUE  \citep{wang2019superglue} tasks given just 32 labeled examples. This is achieved through \emph{priming}: \gpt{} is given a few demonstrations of inputs and corresponding outputs as context for its predictions, but no gradient updates are performed. While being straightforward to use, this method has two major drawbacks: 
\begin{itemize}
	\setlength\itemsep{0.1em}
	\item It requires a gigantic LM to work well, making it \textbf{unusable in many real-world scenarios} and \textbf{resulting in a large carbon footprint} \citep{strubell-etal-2019-energy}.
	\item It \textbf{does not scale to more than a few examples}
          as the context window of most LMs
is limited to
a few hundred tokens.\footnote{While \gpt{} can process up to 2,048 tokens, this is still not enough to fit $\geq$32 examples for some SuperGLUE tasks.}
\end{itemize}

An alternative to priming is \emph{pattern-exploiting
  training} (\pet{}) \citep{schick2020exploiting}, which
combines the idea of reformulating tasks as cloze questions
with regular gradient-based finetuning. While \pet{}
additionally requires unlabeled data, unlabeled data is much easier to obtain than labeled examples for many real-world applications. Crucially, \pet{} only works when the answers to be predicted by the LM correspond to a single token in its vocabulary; this is a severe limitation as many tasks cannot easily be worded that way. 

In this work, we adapt \pet{} for tasks that require predicting multiple tokens. We then show that in combination with ALBERT \citep{lan2019albert}, \pet{} and its iterative variant (\ipet{}) both outperform \gpt{} on SuperGLUE with 32 training examples, while requiring only 0.1\% of its parameters (Figure~\ref{figure:intro}). Moreover, training with \pet{} can be performed in several hours on a single GPU without requiring expensive hyperparameter optimization. Finally, we show that similar performance can also be achieved without unlabeled data and provide a detailed analysis of the factors contributing to \pet{}'s strong performance: its ability to combine multiple task formulations, its resilience to wordings that are hard to understand, its usage of labeled data, and characteristics of the underlying LM.
Given \pet{}'s ``green'' properties,
we see our work as an important contribution to an
environmentally sound NLP.

\section{Related Work}

Enabling LMs to perform zero-shot learning by providing task descriptions was proposed by \citet{radford2018language} and has been applied to text classification \citep{puri2019zeroshot}, commonsense knowledge mining \citep{davison-etal-2019-commonsense} and argumentative relation classification \citep{opitz2019argumentative}. It is also commonly used for probing the knowledge contained within LMs \cite[\emph{i.a.}]{trinh2018simple,Petroni_2019,talmor2019olmpics,schick2019ota,ettinger2020bert}.

As finding ways to reformulate tasks as cloze questions that are understood well by  LMs is difficult \cite{jiang2019know}, \citet{schick2020exploiting} propose \pet{}, a method that uses knowledge distillation \citep{hinton2015distilling} and self-training \citep[e.g.,][]{scudder1965probability,yarowsky-1995-unsupervised,brin1999extracting,mcclosky-etal-2006-effective} to easily combine several reformulations. Our modified version of \pet{} uses masked language models \citep{devlin2018bert} to assign probabilities to sequences of text; this is similar to using them in a generative fashion \cite{wang2019bert} and has previously been investigated by \citet{salazar2019masked} and \citet{ghazvininejad2019maskpredict}. In contrast to \pet{}, which uses gradient-based optimization, \citet{radford2018language} and \citet{brown2020language} investigate priming, where examples are given as context but no parameter updates are performed.

Finally, our focus on reducing the amount of compute required for few-shot learning is closely related to other efforts in Green AI \citep{schwartz2020green} that aim to improve model efficiency, including techniques for knowledge distillation \citep[e.g.,][]{hinton2015distilling,sanh2020distilbert,jiao-etal-2020-tinybert,mao-etal-2020-ladabert,anderson-gomez-rodriguez-2020-distilling}, pruning \citep{NIPS2015_ae0eb3ee,han2015deep_compression,NEURIPS2020_eae15aab} and quantization \citep{gong2014compressing,Zafrir2019Q8BERTQ8,stock2021training} as well as early exit strategies for inference \citep{liu-etal-2020-fastbert,schwartz-etal-2020-right,xin-etal-2020-early}.

\section{Pattern-Exploiting Training}
\label{section:pet}

Let $M$ be a masked language model (MLM),
$T$ its vocabulary and  $\mask{} \in T$ the mask token; we denote
the set of all token sequences as $T^*$. For some $\mathbf{z} \in T^*$ containing at least $k$ masks and $t \in T$, we denote with $q_M^k(t \mid \textbf{z})$ the probability that $M$ assigns to $t$ at the $k$th masked position in $\textbf{z}$; the model's logits before applying softmax are denoted with $s_M^k(t \mid \textbf{z})$.
We consider the task of mapping inputs $x \in X$ to outputs $y \in Y$,
for which \pet{} requires a
set of \emph{pattern-verbalizer pairs} (PVPs). Each PVP $\mathbf{p} = (P, v)$ consists of 
\begin{itemize}
	\item a \emph{pattern} $P: X \rightarrow T^*$ that maps inputs to cloze questions containing a single mask;
	\item a \emph{verbalizer} $v: Y \rightarrow T$ that
          maps each output to a single token representing
          its task-specific meaning in the pattern.
\end{itemize}

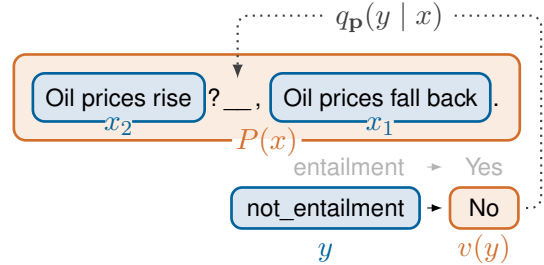
\begin{figure}
	\tikzset{
		every node/.style={
			outer sep=0, text height=1.5ex, text depth=0.25ex
		},
		input/.style={
			draw=c0, rounded corners, line width=2pt
		},
		pattern/.style={
			draw=c1, rounded corners, line width=2pt
		},
		label/.style={
			font=\sffamily\small, rounded corners, inner ysep=0.12cm, inner xsep=0.2cm, outer xsep=0.1cm, text=darkgrey, line width=1pt
		},
		arrow/.style={
			draw=darkgrey,->,>=latex
		},
	}
	\centering
	\begin{tikzpicture}
	
	\path[input] node[partialbox, font=\sffamily\small, fill=c0!10, outer sep=0, inner sep=0.15cm, thick, align=center](input-x1) {Oil prices rise};
	
	\node[font=\sffamily\small, right=0.05cm of input-x1, inner sep=0, outer sep=0](pattern-text-1){?\vphantom{pt}};
	\node[font=\sffamily\small, right=0.05cm of pattern-text-1, inner sep=0, outer ysep=0.1cm](pattern-text-2){\mask{}\vphantom{pt}};
	\node[font=\sffamily\small, right=0.05cm of pattern-text-2, inner sep=0, outer ysep=0.1cm](pattern-text-3){,\ \vphantom{pt}};
	
	\path[input] node[partialbox, font=\sffamily\small, fill=c0!10, outer sep=0, inner sep=0.15cm, thick, align=center, right=0.05cm of pattern-text-3](input-x2) {\textsf{Oil prices fall back}};
	
	\node[font=\sffamily\small, right=0.05cm of input-x2, inner sep=0, outer ysep=0.1cm](pattern-text-3){.\vphantom{pt}};
	
	\node[below=0.025cm of input-x1.south, anchor=center, outer sep=0cm, inner sep=0cm, text=c0](input-label){ ${x}_2$};
	\node[below=0.025cm of input-x2.south, anchor=center, outer sep=0cm, inner sep=0cm, text=c0](input-label){ ${x}_1$};

	\begin{pgfonlayer}{bg}
	\path[pattern] node[partialbox=13pt, fit=(input-x1)(pattern-text-1)(pattern-text-2)(pattern-text-3)(input-x2), fill=c1!10, inner ysep=0.25cm, inner xsep=0.25cm](pattern){};
	\node[below=0.025cm of pattern.south, anchor=center, outer sep=0cm, inner sep=0cm, text=c1](pattern-label){ $P({x})$};
	\end{pgfonlayer}
	
	\node[label, below=0.1cm of pattern.south east, anchor=north east, minimum width=0.9cm, xshift=0.1cm](verbalizer-e){Yes};
	\node[label, below=0cm of verbalizer-e.south west, anchor=north west, text=black, minimum width=0.9cm, fill=c1!10, draw=c1](verbalizer-c){No};
	
	\node[label, left=0.2cm of verbalizer-e](label-e){entailment};
	\node[label, left=0.2cm of verbalizer-c, text=black, fill=c0!10, draw=c0](label-c){not\_entailment};
	
	\path[] (label-e) edge[arrow] (verbalizer-e);
	\path[] (label-c) edge[arrow, draw=black] (verbalizer-c);
	
	\node[below=0.1cm of label-c, text=c0, inner sep=0](y-label){$y\vphantom{v()}$};
	\node[below=0.1cm of verbalizer-c, text=c1, inner sep=0](y-label){$v(y)$};
	
	\draw [black!75, dotted, thick, rounded corners, ->, >=latex] (verbalizer-c.east)--([xshift=0.2cm]verbalizer-c.east)--([xshift=0.2cm, yshift=2.55cm]verbalizer-c.east) -- ([yshift=2.55cm]verbalizer-c.east -| pattern-text-2.center) node [midway, fill=white] {$q_{\mathbf{p}}(y \mid {x})$} -- (pattern-text-2.north);
	
	\end{tikzpicture}
	\caption{Application of a PVP ${\mathbf{p} = (P,v)}$ for recognizing textual entailment: An input ${{x} = ({x}_1, {x}_2)}$ is converted into a cloze question $P({x})$; $q_\mathbf{p}(y \mid {x})$ for each $y$ is derived from the probability of $v(y)$ being a plausible choice for the masked position.}
	\label{figure:pet}
\end{figure}

As illustrated in Figure~\ref{figure:pet}, the core idea of \pet{} is to derive the probability of $y$ being the correct output for ${x}$ from the probability of $v(y)$ being the ``correct'' token at the masked position in $P({x})$.
Based on this intuition, a conditional probability distribution $q_\mathbf{p}$ of $y$ given 
%$X$
$x$
is defined as 
\begin{equation}
q_\mathbf{p}(y \mid {x}) =  \frac{\exp s_\mathbf{p}(y \mid x)}{ \sum_{y' \in Y} \exp s_\mathbf{p}(y' \mid x)} \label{eq:q_p}
\end{equation}
where $s_\mathbf{p}(y \mid x) = s_M^1(v(y) \mid P(x))$ is the raw score of $v(y)$ at the masked position in $P(x)$.

For a given task, identifying PVPs that perform well is challenging in the absence of a large development set. Therefore, \pet{} enables a combination of multiple PVPs $\mathbf{P} = \{ \mathbf{p}_1, \ldots, \mathbf{p}_n \}$ as follows: 
\begin{enumerate}
	\item For each PVP $\mathbf{p}$, a MLM is finetuned on training examples $(x, y)$ by minimizing the cross entropy between $y$ and $q_\mathbf{p}(y \mid x)$. In practice, \citet{schick2020exploiting} train three MLMs per pattern as performance can vary substantially between runs.
	%depending on the order in which training data is presented.
	\item The ensemble of finetuned MLMs is used to annotate a set of unlabeled examples; each unlabeled example $x \in X$ is annotated with soft labels based on the probability distribution 
	\begin{equation}
	q_{\mathbf{P}}(y \mid x) \propto \exp \sum_{\mathbf{p} \in \mathbf{P}} w_\mathbf{p} \cdot s_\mathbf{p}(y \mid x) \label{eq:q_P}
	\end{equation}
	similar to Eq.~\ref{eq:q_p} where $w_\mathbf{p}$ is a weighting term that is proportional to the accuracy achieved with $\mathbf{p}$ on the training set \emph{before} training.
	\item The resulting soft-labeled dataset is used to train a regular sequence classifier by minimizing cross entropy between its output and $q_\mathbf{P}$.
\end{enumerate}
As steps (2) and (3) above closely resemble knowledge distillation \citep{hinton2015distilling}, we also refer to them simply as \emph{distillation}. 
%; both steps are illustrated in Figure~\ref{figure:pet-distillation}.
Importantly, this process does not require holding the entire ensemble of MLMs in memory at the same time as each model's predictions can be computed sequentially; therefore, it is not more memory expensive than using a single model.

To give MLMs trained on different patterns further opportunity to learn from one another, \citet{schick2020exploiting} also propose \ipet{}, an iterative variant of \pet{} in which several generations of models are trained on datasets of increasing size that are labeled by previous generations. 
This is achieved as follows: First, an ensemble of MLMs is trained as in regular \pet{}. For each model $M_i$, a random subset of other models is used to generate a new training set $T_i$ by assigning labels to those unlabeled examples for which the selected subset of models is most confident in its prediction. Each $M_i$ is then retrained on $T_i$; this process is repeated several times, each time increasing the number of examples in $T_i$ by a constant factor. For further details, we refer to \citet{schick2020exploiting}.

\subsection{\pet{} with Multiple Masks}
\label{section:pet-mm}

An important limitation of \pet{} is that the verbalizer $v$ must map each output to a \emph{single} token, which is impossible for many tasks. We thus generalize verbalizers to functions $v: Y \rightarrow T^*$; this requires some modifications to inference and training.\footnote{While \pet{} can easily be adapted to generative MLMs \citep[e.g.,][]{lewis2019bart,raffel2019exploring}, we stick with regular MLMs as they are more lightweight and performed better on simple cloze tasks in preliminary experiments.}
We further generalize \pet{} in that we do not assume the output space to be identical for each input: for each $x \in X$, we denote with $Y_x \subseteq Y$ the set of possible outputs given $x$ as input. Given a PVP $\mathbf{p} = (P, v)$, we define $l(x) = \max_{y \in Y_x}|v(y)|$ to be the maximum number of tokens required to express any output in $Y_x$ and $P^k(x)$ to be $P(x)$ with the mask token replaced by $k$ masks.

\begin{figure}
	\tikzset{
		every node/.style={
			outer sep=0, text height=1.5ex, text depth=0.25ex
		},
		input/.style={
			draw=c0, rounded corners, line width=2pt
		},
		pattern/.style={
			draw=c1, rounded corners, line width=2pt
		},
		label/.style={
			font=\sffamily\small, rounded corners, inner ysep=0.12cm, inner xsep=0.2cm, outer xsep=0.1cm, text=darkgrey, line width=1pt
		},
		arrow/.style={
			draw=black!75,->,>=latex, dotted, thick
		},
	}
	\centering
	\begin{tikzpicture}
	
	\path[input] node[partialbox, font=\sffamily\small, fill=c0!10, outer sep=0, inner sep=0.15cm, thick, align=center](input-x1) {Awful pizza!};
	
	\node[font=\sffamily\small, right=0.15cm of input-x1, inner sep=0, outer sep=0](pattern-text-1){It was \vphantom{pt}};
	\node[font=\sffamily\small, right=0.15cm of pattern-text-1, inner sep=0, outer ysep=0.1cm](pattern-text-2){\mask{}\vphantom{pt}};
	\node[font=\sffamily\small, right=0.15cm of pattern-text-2, inner sep=0, outer ysep=0.1cm](pattern-text-3){\mask{}\vphantom{pt}\negphantom{\mask{}}\phantom{$\sbullet$ble}};
	\node[font=\sffamily\small, right=0.05cm of pattern-text-3, inner sep=0, outer ysep=0.1cm](pattern-text-4){.\vphantom{pt}};
	
	\node[below=0.025cm of input-x1.south, anchor=center, outer sep=0cm, inner sep=0cm, text=c0](input-label){${x}$};
	
	\node[below=0.7cm of pattern-text-2, xshift=-1cm, inner xsep=0, outer xsep=0](terri){$q_M^1(\text{\sffamily\small terri}\,{\mid}\,\mathbf{z})$};
	\node[right=0.1cm of terri, inner xsep=0, outer xsep=0](less){\textcolor{c1}{$\pmb{<}$}};
		\node[right=0.1cm of less, inner xsep=0, outer xsep=0](ble){$q_M^2(\sbullet\text{\sffamily\small ble}\,{\mid}\, \mathbf{z})$};
	
	%\draw [black!75, dotted, thick, rounded corners, ->, >=latex] (terri)--(pattern-text-2.south);
	\path[] (terri.north) edge[bend right=20, dotted, thick, black!75, ->, >=latex] node [left] {} (pattern-text-2.south);
	\path[] (ble.north) edge[bend left=10, dotted, thick, black!75, ->, >=latex] node [left] {} (pattern-text-3.south);
	
	\begin{pgfonlayer}{bg}
	\path[pattern] node[partialbox=17pt, fit=(input-x1)(pattern-text-1)(pattern-text-2)(pattern-text-3)(pattern-text-4), fill=c1!10, inner ysep=0.25cm, inner xsep=0.25cm](pattern){};
	\node[below=0.025cm of pattern.south, anchor=center, outer sep=0cm, inner sep=0cm, text=c1](pattern-label){ $P^2({x})$};
	\end{pgfonlayer}
	
	\node[left=0cm of pattern](){(a)$\quad\mathbf{z}\phantom{'}\,{=}$};
	
	\path[input] node[below=1.8cm of input-x1, partialbox, font=\sffamily\small, fill=c0!10, outer sep=0, inner sep=0.15cm, thick, align=center](input-x1b) {Awful pizza!};
	
	\node[font=\sffamily\small, right=0.15cm of input-x1b, inner sep=0, outer sep=0](pattern-text-1b){It was \vphantom{pt}};
	\node[font=\sffamily\small, right=0.15cm of pattern-text-1b, inner sep=0, outer ysep=0.1cm](pattern-text-2b){\mask{}\vphantom{pt}};
	\node[font=\sffamily\small, right=0.15cm of pattern-text-2b, inner sep=0, outer ysep=0.1cm](pattern-text-3b){$\sbullet$ble\vphantom{pt}};
	\node[font=\sffamily\small, right=0.05cm of pattern-text-3b, inner sep=0, outer ysep=0.1cm](pattern-text-4b){.\vphantom{pt}};
	
	\node[below=0.025cm of input-x1b.south, anchor=center, outer sep=0cm, inner sep=0cm, text=c0](input-labelb){${x}$};
	
	\node[below=0.7cm of pattern-text-2b](terrib){$q_M^1(\text{\sffamily\small terri}\,{\mid}\,\mathbf{z}')$};
	\draw [black!75, dotted, thick, rounded corners, ->, >=latex] (terrib)--(pattern-text-2b.south);
	
	\begin{pgfonlayer}{bg}
	\path[pattern] node[partialbox=0pt, fit=(input-x1b)(pattern-text-1b)(pattern-text-2b)(pattern-text-3b)(pattern-text-4b), fill=c1!10, inner ysep=0.25cm, inner xsep=0.25cm](patternb){};
	\end{pgfonlayer}
	
	\node[left=0cm of patternb](){(b)$\quad\mathbf{z}'\,{=}$};
	
	\end{tikzpicture}
	\caption{Inference for a verbalization consisting of the two tokens \textsf{\small{terri}} and $\sbullet$\textsf{\small{ble}}. (a) We first compute the probability of each token at its position in the cloze question $P^2(x)$ and identify the token with the highest probability. (b) We insert this token into the cloze question and compute the probability of the remaining token.}
	\label{figure:pet-mm}
\end{figure}
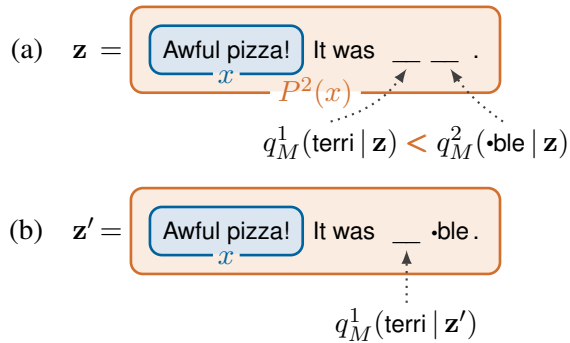

As a running example, we consider the task of binary sentiment classification for restaurant reviews with labels $Y = \{+1, -1\}$. We use the pattern 
$
P(x) = \inlinepattern{$x$\textsf{\small{. It was \mask{} .}}}
$
and a verbalizer $v$ that maps $+1$ to the single token \textsf{\small great} and $-1$ to the sequence \textsf{\small terri} \textsf{\small $\sbullet$ble}, i.e., we assume that the MLM's tokenizer splits the word ``terrible'' into the two tokens \textsf{\small terri} and \textsf{\small $\sbullet$ble}. For this example, $l(x) = 2$ for all $x$; $P^2(x)$ is illustrated in Figure~\ref{figure:pet-mm} (a).

\paragraph{Inference}

For $x \in X$, $y \in Y_x$ and $|v(y)|=k$, we redefine $q_\mathbf{p}(y \mid x)$ in an autoregressive fashion: Starting from $P^k(x)$, we perform $k$ consecutive predictions, where we always select the next token to predict based on the MLM's confidence. That is, we set $q_\mathbf{p}(y \mid x) = q(v(y) \mid P^k(x))$ where
\begin{equation}
q(t_1 \compellipsis t_k {\mid} \mathbf{z}) = 
	\begin{cases} 
		1&\hskip-5pt\text{if } k\,{=}\,0 \\
		q_M^j(t_j {\mid} \mathbf{z})\,{\cdot}\,q(t' {\mid} \mathbf{z}')&\hskip-5pt \text{if } k\,{\geq}\,1
	\end{cases}
\label{eq:q-multimask}
\end{equation}
with $j = \argmax_{i=1}^k q_M^i(t_i \mid \mathbf{z})$, $\mathbf{z}'$ is $\mathbf{z}$ except $\mathbf{z}'_j = t_j$ and  $t' = t_1\compellipsis t_{j-1} t_{j+1} \compellipsis t_k$. Note that unlike in original \pet{} (Eq.~\ref{eq:q_p}), $q_\mathbf{p}$ is not a probability distribution as its values do not sum to one.

For our sentiment classification example, Figure~\ref{figure:pet-mm} illustrates how $q_\mathbf{p}(-1 \mid x)$ is computed: As $|v(y)| = |\{ \textsf{\small{terri}}, \sbullet\textsf{\small{ble}} \}| = 2$, we first use $\mathbf{z} = P^2(x)$ to compute the probability of each token in $v(y)$ (Figure~\ref{figure:pet-mm}a). We then choose the token with the highest probability, put it in place of the corresponding mask token, and use the resulting cloze question $\mathbf{z'}$ to compute the probability of the remaining token (Figure~\ref{figure:pet-mm}b). The overall score for $y = -1$ is then computed as \[
q_\mathbf{p}(-1 \mid x ) = q_M^2(\sbullet\textsf{\small{ble}} \mid \mathbf{z}) \cdot q_M^1(\textsf{\small{terri}} \mid \mathbf{z}')
\]

\paragraph{Training}

Computing $q_\mathbf{p}(y \mid x)$ as in Eq.~\ref{eq:q-multimask} for each training example $(x, y)$ would be prohibitively expensive. To enable computation of all required probabilities in a single forward pass, we approximate $q_\mathbf{p}(y \mid x)$ by (i) always inserting the maximum number of mask tokens required to express any output and (ii) for each $y' \in Y_x$, predicting all tokens in $v(y') = t_1 \ldots t_k$ in parallel, where we simply ignore the model's predictions for all $l(x)-k$ superfluous mask tokens:
\begin{equation}
\tilde{q}_\mathbf{p}(y' \mid x) = \prod_{i=1}^{k} q_M^i(t_i \mid P^{l(x)}(x)) \label{eq:q-tilde}
\end{equation}
For our running example, this means we approximate the scores $q_\mathbf{p}(y \mid x)$ by computing
\begin{align*}
\tilde{q}_\mathbf{p}(+1 \mid x) & = q_M^1 (\textsf{\small{great}} \mid \mathbf{z}) \\
\tilde{q}_\mathbf{p}(-1 \mid x) & = q_M^1 (\textsf{\small{terri}} \mid \mathbf{z}) \cdot q_M^2(\sbullet\textsf{\small{ble}} \mid \mathbf{z})
\end{align*}
which can be done in a single forward pass as it only requires processing the cloze question $\mathbf{z} = P^2(x)$ shown in Figure~\ref{figure:pet-mm} (a) once.

%Based on these approximations, we design the training loss following the principle of PET to update the LM's parameters as little as possible. 
As $\tilde{q}_\mathbf{p}$ is not a probability distribution over $Y_x$, cross entropy is not an ideal training objective as it can also be minimized by reducing the probability assigned to sequences $\mathbf{z} \notin v(Y_x)$ that are not part of the output space, despite this having no effect on the model's prediction.
%; in contrast, changing the probability of any $t \notin v(Y)$ has no effect on the loss in original \textsc{Pet}. 
We instead opt for multi-class hinge loss \citep{weston1999support,dogan2016unified} and minimize:
\begin{equation}
\sum_{y' \in Y_x} \text{max}\left(0;1 {-} \log \tilde{q}_\mathbf{p}(y {\mid} x){+}\log \tilde{q}_\mathbf{p}(y' {\mid} x)\right)
\end{equation}
That is, we require the difference between the log probability of $y$ and the log probability of any output $y' \in Y_x \setminus \{ y \}$ to be at least $1$.

\section{Experiments}
\label{section:experiments}
 
We compare \pet{} and \gpt{}
on SuperGLUE  \cite{wang2019superglue}, a natural language understanding benchmark consisting of eight challenging tasks.
We cannot evaluate \pet{} using the exact same training data as \gpt{} because for most tasks, \gpt{} uses a different set of training examples for each test example and for the other tasks, training sets were not available upon request; however, the exact choice of examples has little impact on \gpt{}'s performance.\footnote{Based on personal correspondence with the authors.} We thus create new training sets by randomly selecting 32 examples for each task using a fixed random seed.

We additionally create sets of up to 20,000 unlabeled examples for each task; this is done by removing all labels from the original training sets. We refer to the resulting sets of training examples and unlabeled examples as \emph{FewGLUE}.\footnote{FewGLUE is publicly available at \url{https://github.com/timoschick/fewglue}.}

\subsection{Tasks}
\label{subsection:tasks}

Below, we describe each of the SuperGLUE tasks and our corresponding PVPs. We use a vertical bar ($|$) to mark
boundaries between text segments. Of the eight tasks considered, only COPA, WSC and ReCoRD require the use of \pet{} with multiple masks as introduced in Section~\ref{section:pet-mm}. \\[-0.5em]

\noindent \textbf{BoolQ} \citep{clark2019boolq} is a QA task where each example consists of a passage $p$ and a yes/no question $q$. We use the following patterns:
\begin{itemize}[topsep=0.5em]
	\setlength\itemsep{-0.1em}
	\item \pattern{$p$\textsf{\small. Question: }$q$\textsf{\small? Answer: \mask{}.}}
	\item \pattern{$p$\textsf{\small. Based on the previous passage, }$q$\textsf{\small? \mask{}.}}
	\item \pattern{\textsf{\small Based on the following passage, }$q$\textsf{\small? \mask{}. }$p$}
\end{itemize}
We define two verbalizers mapping questions containing a true statement to \textsf{\small yes}/\textsf{\small true} and others to \textsf{\small no}/\textsf{\small false}, respectively, for a total of 6 PVPs. \\[-0.5em]

{
\setlength{\abovedisplayskip}{0.5em}
\setlength{\belowdisplayskip}{0.5em}
\noindent \textbf{CB} \citep{demarneffe:cb} and \textbf{RTE} \citep{dagan2006pascal} are textual entailment tasks
like MNLI, so we use PVPs similar to
\citet{schick2020exploiting}. For a premise $p$ and
hypothesis $h$, we use 
	\[
	\pattern{$h$\textsf{\small?$\,|\,$\mask{}, }$p$},
	\pattern{\textsf{\small``}$h$\textsf{\small''?$\,|\,$\mask{}, ``}$p$\textsf{\small''}},
	\pattern{$h$\textsf{\small?$\,|\,$\mask{}. }$p$},
	\pattern{\textsf{\small``}$h$\textsf{\small''?$\,|\,$\mask{}. ``}$p$\textsf{\small''}}
	\]
and a verbalizer that maps entailment to \textsf{\small yes}, disagreement to \textsf{\small no} and neutral to \textsf{\small maybe}.\\[-0.5em]

\noindent Given a premise $p$, the task in \textbf{COPA} \citep{roemmele2011choice} is to determine the \emph{cause} or \emph{effect} of the premise given two options $c_1$ and $c_2$. For determining the \emph{effect}, we use the following patterns:
	\[
	\pattern{\textsf{\small``}$c_1$\textsf{\small'' or ``}$c_2$\textsf{\small''? }$p$\textsf{\small, so \mask{}.}},
	\pattern{$c_1$\textsf{\small\ or }$c_2$\textsf{\small? }$p$\textsf{\small, so \mask{}.}}
	\]
For determining the \emph{cause}, we use the same patterns but replace \textsf{\small so} with \textsf{\small because}. The verbalizer for $c_1$ and $c_2$ is the identity function.}\\[-0.5em]

%\noindent Like CB, \textbf{RTE} \citep{dagan2006pascal} is a textual entailment
%task, so we use the same PVPs. \\[-0.5em]

\begin{table*}
	\small
	\centering
	\setlength\tabcolsep{0.6em}
	\begin{tabularx}{\linewidth}{lXrccccccccc}
		\toprule%----------------------------------------------------------------------------------------------------------------------------------------------------------------
		&              & \bt Params & \bt BoolQ & \bt CB                & \bt COPA & \bt RTE  & \bt WiC  & \bt WSC  & \bt MultiRC           & \bt ReCoRD            & \bt Avg  \\
		& \bt Model    & (M)        & Acc.      & Acc. / F1             & Acc.     & Acc.     & Acc.     & Acc.     & EM / F1a              & Acc. / F1             & --       \\
		\midrule%----------------------------------------------------------------------------------------------------------------------------------------------------------------
		\multirow{10}{*}{\rotatebox[origin=c]{90}{dev}} 
		& \gpt{} Small & 125        & 43.1      & 42.9 / 26.1           & 67.0     & 52.3     & 49.8     & 58.7     & \pzero6.1 / 45.0      & 69.8 / 70.7           & 50.1     \\
		& \gpt{} Med   & 350        & 60.6      & 58.9 / 40.4           & 64.0     & 48.4     & 55.0     & 60.6     & 11.8 / 55.9           & 77.2 / 77.9           & 56.2     \\
		& \gpt{} Large & 760        & 62.0      & 53.6 / 32.6           & 72.0     & 46.9     & 53.0     & 54.8     & 16.8 / 64.2           & 81.3 / 82.1           & 56.8     \\
		& \gpt{} XL    & 1,300      & 64.1      & 69.6 / 48.3           & 77.0     & 50.9     & 53.0     & 49.0     & 20.8 / 65.4           & 83.1 / 84.0           & 60.0     \\
		& \gpt{} 2.7B  & 2,700      & 70.3      & 67.9 / 45.7           & 83.0     & 56.3     & 51.6     & 62.5     & 24.7 / 69.5           & 86.6 / 87.5           & 64.3     \\
		& \gpt{} 6.7B  & 6,700      & 70.0      & 60.7 / 44.6           & 83.0     & 49.5     & 53.1     & 67.3     & 23.8 / 66.4           & 87.9 / 88.8           & 63.6     \\
		& \gpt{} 13B   & 13,000     & 70.2      & 66.1 / 46.0           & 86.0     & 60.6     & 51.1     & 75.0     & 25.0 / 69.3           & 88.9 / 89.8           & 66.9     \\
		& \gpt{}       & 175,000    & 77.5      & 82.1 / 57.2           & 92.0     & 72.9     & \bt 55.3 & 75.0     & 32.5 / 74.8           & {\bt 89.0} / \bt 90.1 & 73.2     \\
		& \pet{}       & 223        & 79.4      & 85.1 / 59.4           & \bt 95.0 & 69.8     & 52.4     & \bt 80.1 & {\bt 37.9} / \bt 77.3 & 86.0 / 86.5           & 74.1     \\
		& \ipet{}      & 223        & \bt 80.6  & {\bt 92.9} / \bt 92.4 & \bt 95.0 & \bt 74.0 & 52.2     & \bt 80.1 & 33.0 / 74.0           & 86.0 / 86.5           & \bt 76.8 \\
		\midrule%----------------------------------------------------------------------------------------------------------------------------------------------------------------
		\multirow{4}{*}{\rotatebox[origin=c]{90}{test}}     
		& \gpt{}       & 175,000    & 76.4      & 75.6 / 52.0           & \bt 92.0 & 69.0     & 49.4     & 80.1     & 30.5 / 75.4           & {\bt 90.2} / \bt 91.1 & 71.8     \\
		& \pet{}       & 223        & 79.1      & 87.2 / 60.2           & 90.8     & 67.2     & \bt 50.7 & \bt 88.4 & {\bt 36.4} / \bt 76.6 & 85.4 / 85.9           & 74.0     \\
		& \ipet{}      & 223        & \bt 81.2  & {\bt 88.8} / \bt 79.9 & 90.8     & \bt 70.8 & 49.3     & \bt 88.4 & 31.7 / 74.1           & 85.4 / 85.9           & \bt 75.4 \\
		& SotA         & 11,000     & \et 91.2  & {\et 93.9} / \et 96.8 & \et 94.8 & \et 92.5 & \et 76.9 & \et 93.8 & {\et 88.1} / \et 63.3 & {\et 94.1} / \et 93.4 & \et 89.3 \\
		\bottomrule%-------------------------------------------------------------------------------------------------------------------------------------------------------------
	\end{tabularx}
	\caption{Results on SuperGLUE for \gpt{} primed with
		32 randomly selected examples and for \pet{} /
		\ipet{} with ALBERT-xxlarge-v2 after training on
		FewGLUE. State-of-the-art results when using the
		regular, full size training sets for all tasks \citep{raffel2019exploring} are shown in italics.
	}
	\label{table:main_results}
\end{table*}

\noindent For \textbf{WiC} \citep{pilehvar2018wic}, given a word $w$ and two sentences $s_1$ and $s_2$ in which it occurs, the task is to decide if $w$ is used with the same sense in both sentences. We use:
\begin{itemize}[topsep=0.5em]
	\setlength\itemsep{-0.1em}
	\item \pattern{\textsf{\small``}$s_1$\textsf{\small'' / ``}$s_2$\textsf{\small''. Similar sense of ``}$w$\textsf{\small''? \mask{}.}}
	\item 
		\begin{multipattern}
		$s_1\ s_2$\textsf{\small\ Does }$w$\textsf{\small\ have the same meaning in both sentences? \mask{}}
		\end{multipattern}
	\item \pattern{$w$\textsf{\small. Sense (1) (a) ``}$s_1$\textsf{\small'' (\mask{}) ``}$s_2$\textsf{\small''}}
\end{itemize}
For the first two patterns, we use \textsf{\small yes} as verbalization for words used in the same sense and \textsf{\small no} for other words; for the third pattern, we use \textsf{\small b} and \textsf{\small 2}.\\[-0.5em]

\noindent For \textbf{WSC} \citep{levesque2011winograd}, each example consists of a sentence $s$ with a marked pronoun $p$
and noun $n$, and the task is to determine whether $p$ refers to $n$. We follow  \citep{raffel2019exploring,brown2020language} and treat WSC as a generative task. We highlight $p$ in $s$ by putting it in asterisks and use the following patterns:
\begin{itemize}[topsep=0.5em]
	\setlength\itemsep{-0.1em}
	\item \pattern{$s$\textsf{\small\ The pronoun `}$*p*$\textsf{\small' refers to \mask{}.}}
	\item 
		\begin{multipattern}
		$s$\textsf{\small\ In the previous sentence, the pronoun `}$*p*$\textsf{\small' refers to \mask{}.}
		\end{multipattern}
	\item 
		\begin{multipattern}
		$s$\textsf{\small\ In the passage above, what does the pronoun `}$*p*$\textsf{\small' refer to? Answer: \mask{}.}
		\end{multipattern}
\end{itemize}
We use the identity function as verbalizer for $n$. Note that WSC is different from other tasks in that it requires free-form completion. This in turn requires some modifications during training and inference that are discussed in Appendix~\ref{appendix:training_details}.\\[-0.5em]

\noindent \textbf{MultiRC} \citep{khashabi2018looking} is a QA task. Given a passage $p$, a question $q$ and an answer candidate $a$, the task is to decide whether $a$ is a correct answer for $q$. We use the same verbalizer as for BoolQ and similar patterns:
\begin{itemize}[topsep=0.5em]
	\setlength\itemsep{-0.1em}
	\item \pattern{$p$\textsf{\small. Question: }$q$\textsf{\small? Is it }$a$\textsf{\small? \mask{}.}}
	\item \pattern{$p$\textsf{\small. Question: }$q$\textsf{\small? Is the correct answer ``}$a$\textsf{\small''? \mask{}.}}
	\item 
		\begin{multipattern}
		$p$\textsf{\small. Based on the previous passage, }$q$\textsf{\small? Is ``}$a$\textsf{\small'' a correct answer? \mask{}.}
		\end{multipattern}
\end{itemize} \vspace{0.5em}

\noindent For \textbf{ReCoRD} \citep{zhang2018record}, given a passage $p$ and a cloze question $q$, the task is to decide which of a given set of answer candidates is the correct replacement for the placeholder in the cloze question. As this task is already presented in the form of a cloze question, there is little room for designing PVPs, so we only use a trivial one: the concatenation of $p$ and $q$ as pattern and the identity function as verbalizer. With only one PVP, there is no need to perform knowledge distillation so we directly use the resulting model as our final classifier.

\subsection{Setup}

As underlying LM for \pet{} we choose
ALBERT-xxlarge-v2 \citep{lan2019albert}, the best-performing
MLM on SuperGLUE when training is performed on the regular,
full size training sets. We use the same model, supplemented by a sequence classification head, as our final classifier. We run
\pet{} on the FewGLUE training sets for all SuperGLUE tasks. We do not use any development set to optimize hyperparameters; instead we use the exact same setup and hyperparameters as
\citet{schick2020exploiting}. For COPA, WSC and ReCoRD, we
use our proposed modification of \pet{} to support
verbalizers mapping labels to multiple tokens; for all other
tasks, we use regular \textsc{Pet}. We train \ipet{} on all
tasks except COPA and WSC, as their unlabeled sets contain
well below 1,000 examples, as well as ReCoRD, for which \ipet{} makes no sense as we only use a single PVP. For these three tasks, we simply reuse the results of regular \pet{}.

\subsection{Results}

Our main results are shown in
Table~\ref{table:main_results}. As can be seen, ALBERT with
\pet{} performs similar to the largest \gpt{} model, which
is larger by a factor of 785. On average, \pet{} performs 18
points better compared to \gpt{} Med, a model of similar size. \ipet{} brings further improvements for 3 out of the 5 tasks that we use \ipet{} for, most notably for CB, but results in a slight performance drop for MultiRC. 
%Similar to \gpt{}, we find that \pet{} does not perform above random chance for WiC, which is difficult to reformulate as a language modeling task. 
% ReCoRD is the only task for which \gpt{} consistently performs better than both \pet{} and \ipet{}.
%; however, it is also the only task for which we define just a single PVP and perform no knowledge distillation. 
Despite \pet{}'s strong performance, it still clearly performs worse than a state-of-the-art model trained on the regular, full size SuperGLUE training set.

\section{Analysis}

We investigate the importance of several factors for
few-shot performance: the choice of patterns and
verbalizers, the usage of both unlabeled and labeled data,
and properties of the underlying language model. We also
look into our proposed modification for \pet{} to work with
multiple masks and compare it to various baselines. Finally,
we measure how choosing different sets of training examples
affects performance. Our analysis focuses  on \pet{}
as \gpt{} is not publicly available.\footnote{We
could not obtain access to OpenAI's \gpt{} API.}

\subsection{Patterns}

The way in which tasks are reformulated as cloze questions can have a huge impact on performance \citep{jiang2019know,schick2020exploiting}. These reformulations can be arbitrarily complex; for example, the pattern used by \gpt{} for WSC contains an introductory section of almost 30 words; it is unclear if and how this formulation has been optimized.\footnote{While the authors use a different terminology, \gpt{} also makes use of PVPs \citep[pp.~50--61]{brown2020language}.} To investigate the importance of patterns and verbalizers, we compare three sets of PVPs: our initial set as defined in Section~\ref{subsection:tasks} (denoted $\mathbf{p}_\text{ours}$), the single PVP used by \gpt{} ($\mathbf{p}_\text{\gpt{}}$), and the combination of both ($\mathbf{p}_\text{comb}$).

We train ALBERT using \pet{} with all three sets of patterns; results for selected SuperGLUE tasks are shown in Table~\ref{table:pattern_results} (top). As can be seen, the PVP used by \gpt{} outperforms our PVPs on RTE whereas our initial set of patterns performs much better on MultiRC.
%, with an absolute improvement of 9 points in F1a. 
These large differences in performance highlight the importance of finding good ways to express tasks as cloze questions. As it is difficult to ascertain which patterns perform well without trying them on a large set of examples, a key challenge for few-shot approaches is to compensate for PVPs that the 
%used
LM fails to understand well. As seen in the performance
of the model trained with $\mathbf{p}_\text{comb}$, \pet{}
is able to do so: not only does combining all PVPs
compensate for the worse performance of
$\mathbf{p}_\text{ours}$ on RTE and of
$\mathbf{p}_\text{\gpt{}}$ on MultiRC, it even further
improves average performance across the three tasks
compared to the best-performing set of patterns. This clearly demonstrates the potential of carefully engineering a set of suitable patterns as opposed to just choosing a single formulation without means of evaluating its effectiveness.

\begin{table}
	\small
	\centering
	\setlength\tabcolsep{0.5em}
	\begin{tabularx}{\linewidth}{Xcccc} 
	\toprule%------------------------------------------------------------------------------------------------
	                                         & \bt CB            & \bt RTE  & \bt MultiRC       & \bt Avg  \\
	\bt Model                                & Acc. / F1         & Acc.     & EM / F1a          & --       \\
	\midrule%------------------------------------------------------------------------------------------------
	\pet{} ($\mathbf{p}_\text{ours}$)        & {\bt 85.1} / 59.4 & 69.8     & 37.9 / 77.3       & 66.6     \\
	\pet{} ($\mathbf{p}_\text{\gpt{}}$)      & 83.3 / 58.1       & 71.8     & 25.4 / 68.3       & 63.1     \\
	\pet{} ($\mathbf{p}_\text{comb}$)        & 84.5 / 59.0       & \bt 74.7 & 39.1 / \bt 77.7   & 68.3     \\\midrule
	\pet{} ($\mathbf{p}_\text{ours}$) $\neg$dist & 83.9 / \bt 76.2   & 66.4     & 38.9 / 76.2       & 68.0     \\
	\pet{} ($\mathbf{p}_\text{comb}$) $\neg$dist & 83.9 / \bt 76.2   & 72.9     & {\bt 39.6} / 76.6 & \bt 70.4 \\
	\bottomrule%---------------------------------------------------------------------------------------------
	\end{tabularx}
	\caption{Results on selected tasks for various sets of PVPs  for regular \pet{} and for an ensemble of \pet{} models with no knowledge distillation (``$\neg$dist'')}
\label{table:pattern_results}
\end{table}

\subsection{Unlabeled Data Usage}

Unlike \gpt{}, \pet{} requires unlabeled data to distill the knowledge of all models based on individual PVPs into a single classifier; for \ipet{}, unlabeled data is additionally used to generate training sets for future generations. The underlying assumption is that unlabeled data can easily be obtained, which may not always be the case in real-world settings. We thus investigate the importance of unlabeled data for regular \pet{}. To this end, we compare the performance of the final classifier in \pet{} to that of directly using the ensemble of models corresponding to individual PVPs. While using this ensemble entirely removes the need for unlabeled data, the ensemble for $k$ PVPs is larger than the distilled model by a factor of $3\cdot k$ as we follow the default setting of \pet{} and train three models per PVP. However, even for a large number of PVPs the ensemble is smaller than \gpt{} by two orders of magnitude.

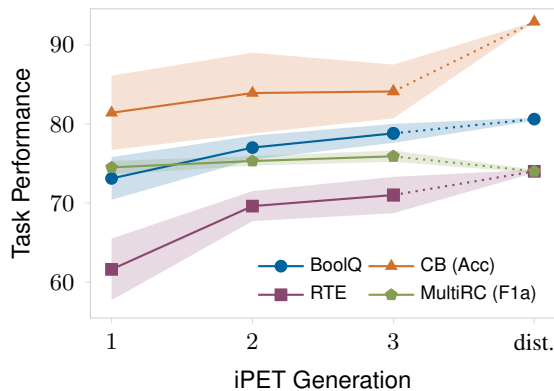
\begin{figure}
	\begin{tikzpicture}
	\begin{axis}[
	cycle list name=color list,
	xlabel={\sffamily\small iPET Generation},
	ylabel={\sffamily\small Task Performance},
	axis line style={decentgrey!95!black},
	major grid style={line width=.2pt,draw=decentgrey},
		enlarge x limits={0.05},
	enlarge y limits={0.075},
	ymin = 58,
	ymax = 92,
	xmin = 1,
	xmax = 4,
	minor tick style={decentgrey!0},
	major tick style={decentgrey},
	xtick pos=left,
	ytick pos=left,
	ylabel near ticks,
	xlabel near ticks,
	xtick={1,2,3,4},
	xticklabels={$1$, $2$, $3$, dist.},
	tick align=outside,
	tick label style={font=\footnotesize},
	major tick length=0.075cm,
	width = \linewidth,
	height = 0.23\textheight,
	log ticks with fixed point,
	x tick label style={/pgf/number format/1000 sep=\,},
	legend style={draw=none, fill=white!75, at={(0.99,0.01)},anchor=south east, font=\sffamily\scriptsize},
	legend cell align=left,
	legend columns=2,
	]
	\addplot[mark=*, c0, thick, mark options={solid}] coordinates {
		(1,73.1)
		(2,77.0)
		(3,78.8)
	};
	\addlegendentry{BoolQ}
	
	\addplot[mark=*, forget plot, c0, thick, dotted, mark options={solid}] coordinates {
		(3,78.8)
		(4,80.6)
	};

    \addplot[name path=boolq-top, opacity=0, forget plot] coordinates {
    	(1,75.8)
    	(2,78.5)
    	(3,80.0)
    	(4,80.8)
	};

	\addplot[name path=boolq-down, opacity=0, forget plot] coordinates {
		(1,70.4)
		(2,75.5)
		(3,77.6)
		(4,80.4)
	};

	\addplot[c0,fill opacity=0.2, forget plot] fill between[of=boolq-top and boolq-down];

	\addplot[mark=triangle*, c1, thick, mark options={solid}] coordinates {
		(1,81.4)
		(2,83.9)
		(3,84.1)
	};
	\addlegendentry{CB (Acc)}
	
	\addplot[mark=triangle*, forget plot, c1, thick, mark options={solid}, dotted] coordinates {
		(3,84.1)
		(4,92.9)
	};
	
	\addplot[name path=cb-top, opacity=0, forget plot] coordinates {
		(1,86.1)
		(2,89.0)
		(3,87.5)
		(4,92.9)
	};
	
	\addplot[name path=cb-down, opacity=0, forget plot] coordinates {
		(1,76.7)
		(2,78.8)
		(3,80.7)
		(4,92.9)
	};
	
	\addplot[c1,fill opacity=0.2, forget plot] fill between[of=cb-top and cb-down];

	\addplot[mark=square*, c2, thick, mark options={solid}] coordinates {
		(1,61.6)
		(2,69.6)
		(3,71.0)
	};
	\addlegendentry{RTE}
	
	\addplot[mark=square*, c2, thick, dotted, forget plot, mark options={solid}] coordinates {
		(3,71.0)
		(4,74.0)
	};

	\addplot[name path=rte-top, opacity=0, forget plot] coordinates {
		(1,65.5)
		(2,71.5)
		(3,73.3)
		(4,74.2)
	};
	
	\addplot[name path=rte-down, opacity=0, forget plot] coordinates {
		(1,57.7)
		(2,67.7)
		(3,68.7)
		(4,73.8)
	};
	
	\addplot[c2,fill opacity=0.2, forget plot] fill between[of=rte-top and rte-down];

	\addplot[mark=pentagon*, c4, thick, mark options={solid}] coordinates {
		(1,74.5)
		(2,75.3)
		(3,75.9)
	};
	\addlegendentry{MultiRC (F1a)}
	
	\addplot[mark=pentagon*, c4, thick, dotted, forget plot, mark options={solid}] coordinates {
		(3,75.9)
		(4,74.0)
	};
	
	\addplot[name path=multirc-top, opacity=0, forget plot] coordinates {
		(1,75.33)
		(2,75.93)
		(3,76.64)
		(4,74.2)
	};
	
	\addplot[name path=multirc-down, opacity=0, forget plot] coordinates {
		(1,73.6)
		(2,74.7)
		(3,75.2)
		(4,73.8)
	};
	
	\addplot[c4,fill opacity=0.2, forget plot] fill between[of=multirc-top and multirc-down];

	\end{axis}
	\end{tikzpicture}
	\caption{Average performance ($\pm$ standard deviation) of all MLMs trained on individual patterns for three generations and of the distilled classifier (``dist.'') across three individual training runs}
	\label{figure:ipet}
\end{figure} 

Results without distillation can be seen in
Table~\ref{table:pattern_results} (bottom). Averaged across
the three tasks, the ensemble performs even better
than the distilled classifier. This shows that
if the goal is only to achieve good performance,
then unlabeled data is not necessary; however, it is required to obtain a single, lightweight model as final classifier. 

Figure~\ref{figure:ipet} illustrates the benefit of training multiple generations with \ipet{}. For all tasks except MultiRC, there are substantial improvements from the first to the second generation, whereas the third generation achieves only slight additional improvements. On average, standard deviation is reduced in later generations, illustrating that the models learn from each other and their predictions converge. The final distillation step brings further improvements for all tasks except MultiRC and reduces standard deviation across three training runs to almost zero, illustrating that \pet{} and \ipet{} are effective means of reducing finetuning instability \citep{dodge2020finetuning}.

Of course, there are further ways to leverage unlabeled data such as keeping an auxiliary language modeling objective during finetuning \cite{chronopoulou-etal-2019-embarrassingly}. While we leave investigating the impact of additionally using such methods to future work, we note that they can easily be applied to \pet{} while there is no straightforward way to combine them with priming.
 
\subsection{Labeled Data Usage}
\label{subsection:labeled_data_usage}

\begin{table}
	\small
	\centering
	\setlength\tabcolsep{0.5em}
	\begin{tabularx}{\linewidth}{Xcccc} 
		\toprule%-----------------------------------------------------------------------------------------
		                & \bt CB                & \bt RTE  & \bt MultiRC                     & \bt Avg  \\
		\bt Model       & Acc. / F1             & Acc.     & EM / F1a                        & --       \\
		\midrule%-----------------------------------------------------------------------------------------
		\pet{}          & {\bt 85.1} / \bt 59.4 & \bt 69.8 & {\bt 37.9} / \bt 77.3           & \bt 66.6 \\
		unsupervised    & 33.5 / 23.1           & 55.0     & \pzero3.9 / 60.3                & 38.5     \\
		supervised      & 60.7 / 42.5           & 50.2     & \pzero4.3 / 49.8                & 43.0     \\
		\midrule%-----------------------------------------------------------------------------------------
		\pet{} (XLNet)  & {\bt 88.7} / \bt 83.0 & \bt 60.4 & {\bt 21.4} / \bt 66.6           & \bt 63.4 \\
		Priming (XLNet) & 56.3 / 37.7           & 49.5     & \pzero--\pzero / \pzero--\pzero & --       \\
		\bottomrule%--------------------------------------------------------------------------------------
	\end{tabularx}
	
	\caption{Results on selected tasks for various ways of using the labeled examples available in FewGLUE}
	\label{table:labeled_results}
\end{table}

We next investigate the effect of how labeled data is used, which is one of the key differences between priming and \pet{}. We first compare \pet{} with regular supervised training (i.e., without using any patterns), and with a fully unsupervised model (i.e., an ensemble using all PVPs but no labeled training examples). Given 32 examples, \pet{} clearly outperforms both baselines (Table~\ref{table:labeled_results}).%, which is in line with findings by \citet{schick2020exploiting}. 

We next compare \pet{} directly to priming. However, we
cannot do so using ALBERT as it %, like most pretrained MLMs,
is only able to process sequences of up to 512 tokens, which
is not enough for a set of 32 examples;
%As it theoretically supports sequences of arbitrary length, 
we instead use XLNet
\citep{NIPS2019_8812} for this comparison. As shown in
Table~\ref{table:labeled_results}, XLNet in general performs
worse than ALBERT. More importantly, XLNet with \pet{}
performs much better than priming.
% (and
%priming  does not even
%perform above random chance for RTE). 
We were not able to
obtain results with priming on MultiRC because the 32
examples in FewGLUE would require more than 10,000 tokens, so
processing them with a standard Transformer
\citep{Vaswani2017} is infeasible due to the quadratic
complexity of self-attention. This highlights another
important issue with priming: It does not scale well to more
than a few examples; even \gpt{} is only able to process
sequences of up to 2,048 tokens. While there are some
Transformer variants that can deal with much longer
contexts
\citep[e.g.,][]{kitaev2020reformer,beltagy2020longformer}, it
has yet to be investigated to what extent such models make good use of priming examples over 
long context spans.

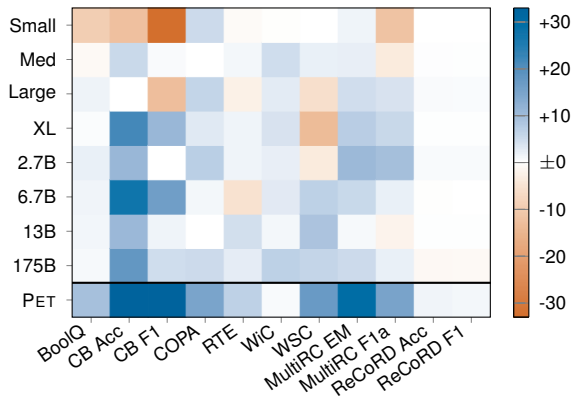
\begin{figure}
	\centering
	\begin{tikzpicture}
	\begin{axis}[
	width=0.92\linewidth,
	height=0.23\textheight,
	view={0}{90},
	colorbar,
	colormap = {blackwhite}{color(0cm)  = (c1);color(0.5cm) = (white);color(1cm) = (c0)},
	colorbar style={
		yticklabel style={
			/pgf/number format/.cd,
			fixed,
			precision=1,
			fixed zerofill,
		},
		width=0.2cm,
		ytick={-30, -20, -10, 0, 10, 20, 30},
		yticklabels={-30, -20, -10,$\pm$0, +10, +20, +30}
	},
	enlargelimits=false,
	axis on top,
	point meta min=-33,
	point meta max=33,
	ymin=-0.5,
	ymax=8.5,
	xmin=-0.5,
	xmax=10.5,
	xtick={0,1,2,3,4,5,6,7,8,9,10},
	xticklabels={BoolQ, CB Acc, CB F1, COPA, RTE, WiC, WSC, MultiRC EM, MultiRC F1a, ReCoRD Acc, ReCoRD F1},
	xticklabel style={rotate=35, anchor=east},
	yticklabels={\pet{}, 175B, 13B, 6.7B, 2.7B, XL, Large, Med, Small},
	ytick={0,1,2,3,4,5,6,7,8,9,10},
	xtick pos=left,
	ytick pos=left,
	ylabel near ticks,
	xlabel near ticks,
	tick align=outside,
	major tick length=0.075cm,
	tick label style={font=\sffamily\scriptsize}
	]
	
	\addplot[matrix plot*,point meta=explicit] table [x=task, y=model, meta=val, col sep=comma] {heatmap.csv};
	
	% add a black line between GPT-3's results and PET results
	\addplot[black, thick, mark options={solid}] coordinates {
		(-1,0.5)
		(15,0.5)
	};
	\end{axis}
	\end{tikzpicture}
	\caption{Accuracy differences between priming with 32 examples and one-shot priming for all \gpt{} models as well as between ALBERT with \pet{} (without distillation) and unsupervised ALBERT (bottom row)}
	\label{figure:heatmap}
\end{figure}

We further investigate the effectiveness of priming by
looking at results obtained with \gpt{} more closely.
%To analyze how well \gpt{} is able to make use of examples given as context, 
To this end, Figure~\ref{figure:heatmap} shows the performance difference
between priming \gpt{} with 32 examples and priming it with just a single
example for each task and model size.\footnote{
	We do not compare priming to zero-shot performance as for unknown reasons,
	zero-shot \gpt{} performs well below
	random guessing for some tasks (e.g., 0.0\% accuracy for WiC).
    To not overestimate the	benefit of priming%
    % with multiple examples
, we therefore
	show gains from providing 32 examples compared
	to just one.} As can be seen, priming with 32
examples only slightly improves performance for most tasks and model sizes. For some tasks, adding more examples even
leads to worse performance, especially for smaller models. For ReCoRD, even the largest model's performance slightly drops when adding more examples.

The bottom row of Figure~\ref{figure:heatmap} shows the performance difference between ALBERT trained with \pet{} (without distillation) and a fully unsupervised ALBERT model on all tasks. While results are not directly comparable due to different underlying models and PVPs, \pet{} results in much stronger performance improvements compared to priming and does not worsen results for any task.

\subsection{Model Type}

We next look into the impact of the underlying LM on
\pet{} by comparing ALBERT with RoBERTa large \citep{liu2019roberta} and GPT-2 medium \citep{radford2018language}. As GPT-2 is a unidirectional model similar to \gpt{}, it can only process patterns where the mask token is the very last token. We therefore use $\mathbf{p}_\text{\gpt{}}$ for CB and RTE; for MultiRC, we stick with our original set of patterns as they already fulfill this requirement. We also do not perform distillation and instead report the ensemble's performance as there is no established way of equipping GPT-2 with a sequence classification head. 

\begin{table}
	\small
	\centering
	\setlength\tabcolsep{0.4em}
	\begin{tabularx}{\linewidth}{Xccccc} 
		\toprule%--------------------------------------------------------------------------------------
		          &            & \bt CB                & \bt RTE  & \bt MultiRC           & \bt Avg  \\
		\bt Model & \bt Params & Acc. / F1             & Acc.     & EM / F1a              & --       \\
		\midrule%--------------------------------------------------------------------------------------
		ALBERT    & 223M       & {\bt 87.5} / \bt 78.7 & \bt 74.7 & {\bt 38.9} / \bt 76.2 & \bt 71.8 \\
		RoBERTa   & 355M       & 85.7 / 77.5           & 62.8     & 23.3 / 70.0           & 63.7     \\
		GPT-2     & 345M       & 73.2 / 73.7           & 47.7     & 12.4 / 57.4           & 52.0     \\
		\bottomrule%-----------------------------------------------------------------------------------
	\end{tabularx}
	\caption{Results on selected tasks for \pet{} without knowledge distillation combined with various LMs using $\mathbf{p}_\text{\gpt{}}$ for CB/RTE and $\mathbf{p}_\text{ours}$ for MultiRC}
	\label{table:model_type_results}
\end{table}

Results for training all three LMs with \pet{} in
Table~\ref{table:model_type_results} show that using ALBERT
as underlying LM is crucial for \pet{}'s strong performance;
exchanging ALBERT with RoBERTa results in an average
performance drop of 8 points. However, RoBERTa still clearly
outperforms \gpt{} 13B, which is larger by two orders of
magnitude. Importantly, \pet{} with GPT-2 performs much
worse than with the two other models.
% -- its performance on RTE is not above random chance. 
As anticipated by 
\citet{brown2020language}, a  reason for this drop in
performance may be that like \gpt{}, GPT-2 is
%a
unidirectional,
%LM,
making tasks that require
comparing two sequences  a 
challenge. However, it is important to note that there are
also other substantial differences between GPT-2 and the
other two models, most notably the pretraining
dataset. Regardless of whether
%its
unidirectionality is
the reason for GPT-2's bad performance, bidirectionality of
the underlying LM is important for \pet{} as it removes the
need for the mask token to be at the very end and thus allows for more flexibility in the creation of patterns.

\subsection{\pet{} with Multiple Masks}

We modified \pet{} to work for outputs that require more
than a single token.
To investigate the impact of this modification, we look at
the three tasks for which this is required: COPA, WSC and
ReCoRD. We compare our decoding strategy of predicting tokens in order of the probability assigned to them, to which we refer as \emph{max-first}, with two
alternatives: decoding  left-to-right (ltr) as is common for
many autoregressive language models, and decoding all tokens
simultaneously (parallel) as is done during
training. Additionally, we compare \pet{} with
untrained ALBERT
to measure the effectiveness of our proposed training loss.

Results are shown in
Table~\ref{table:multimask_results}. \pet{} clearly
outperforms  untrained ALBERT  for the three tasks. Not performing distillation hurts performance for COPA, but leads to slight improvements on WSC; for ReCoRD, we did not perform distillation in the first place as we only use a single PVP. 
Our decoding strategy is clearly superior to parallel decoding except for WSC, for which most predictions consist only of one or two tokens, and performs slightly better than left-to-right decoding.

\begin{table}
	\small
	\centering
	\setlength\tabcolsep{0.45em}
	\begin{tabularx}{\linewidth}{Xcccc} 
		\toprule%----------------------------------------------------------------------------
		                          & \bt COPA & \bt WSC  & \bt ReCoRD            & \bt Avg  \\
		\bt Model                 & Acc.     & Acc.     & Acc. / F1             & --       \\
		\midrule%----------------------------------------------------------------------------
		\pet{}                    & \bt 95.0 & 80.1     & {\bt 86.0} / \bt 86.5 & \bt 87.1 \\
		\pet{} $\neg$dist (max-first) & 90.0     & \bt 80.8 & {\bt 86.0} / \bt 86.5 & 85.7     \\
		\pet{} $\neg$dist (ltr)       & 89.0     & 79.8     & 84.7 / 85.3           & 84.6     \\
		\pet{} $\neg$dist (parallel)  & 77.0     & \bt 80.8 & 82.5 / 83.1           & 80.2     \\
%		unsupervised              & 72.5     & 59.9     & 84.7 / 85.4           & 72.5     \\
		untrained                 & 72.5     & 59.9     & 84.7 / 85.4           & 72.5     \\
		\bottomrule%-------------------------------------------------------------------------
	\end{tabularx}
	\caption{Results on selected tasks for %with % various
		our proposed
          variant of \pet{} as well as other
          decoding strategies and for
          %fully unsupervised model}
          untrained ALBERT}
	\label{table:multimask_results}
\end{table}

\subsection{Training Examples}
Recall that we conduct our experiments with training examples from
FewGLUE, a randomly selected subset of the original SuperGLUE training examples. We used
a fixed random seed $s_0$ to generate FewGLUE. Let $\Sigma_i$ be
the randomly selected subset of SuperGLUE for random seed
$s_i$, so $\Sigma_0 =$ FewGLUE. In this subsection, we create two additional subsets
of SuperGLUE, $\Sigma_1$ and $\Sigma_2$, based on different seeds.
This allows us to investigate how 
different sets of training
examples affect performance. 
To this end,
we run \pet{}
for CB, RTE and MultiRC 
using the three
$\Sigma_i$. To
measure only the effect of varying the training set while
ignoring unlabeled examples, we do not use distillation.

Table~\ref{table:seed_results} shows that for all tasks, changing the set of training examples can result in large performance differences for \pet{}. This highlights the importance of using the same set of examples when comparing different few-shot approaches, which is why we make the particular set of examples in FewGLUE publicly available. However, we note that the average performance of \pet{} is similar to that of \gpt{} for all seeds. 

While our results may seem contrary to the insight that for \gpt{}, the exact choice of examples does not play a major role, we suspect this to be due to the fact that priming benefits much less from training examples than \pet{} (cf. Section~\ref{subsection:labeled_data_usage}); accordingly, the influence of the exact set of training examples on the model's performance is smaller.

\begin{table}
	\small
	\centering
	\setlength\tabcolsep{0.55em}
	\begin{tabularx}{\linewidth}{Xcccc} 
		\toprule%------------------------------------------------------------------------------------------
		                           & \bt CB                & \bt RTE  & \bt MultiRC           & \bt Avg  \\
		\bt Model                  & Acc. / F1             & Acc.     & EM / F1a              & --       \\
		\midrule%------------------------------------------------------------------------------------------
		\gpt{}                     & 82.1 / 57.2           & \bt 72.9 & 32.5 / 74.8           & 65.4     \\
		\pet{} $\neg$dist ($\Sigma_0$) & 83.9 / 76.2           & 66.4     & 38.9 / 76.2           & \bt 68.0 \\
		\pet{} $\neg$dist ($\Sigma_1$) & 82.1 / 57.4           & 61.4     & {\bt 39.2} / \bt 77.9 & 63.2     \\
		\pet{} $\neg$dist ($\Sigma_2$) & {\bt 87.5} / \bt 84.0 & 61.4     & 34.7 / 76.3           & 67.6     \\
		\bottomrule%---------------------------------------------------------------------------------------
	\end{tabularx}
	\caption{Results on selected tasks for \gpt{} and
          for \pet{} using training sets $\Sigma_0$,
          $\Sigma_1$,
           $\Sigma_2$}
	\label{table:seed_results}
\end{table}

%\todo{Add discussion of instability due to random shuffling / dropout as in the 2020 paper by Jesse Dodge et al. Same dataset results for PET with STDEV. For RTE, final classifier with regular PET we have a standard deviation of 0.8, CB of Acc 1.0, F1 0.8 and MultiRC of EM 0.6, F1 0.2 on the original training set $\Sigma_0$}

\section{Conclusion}

We have proposed a simple yet effective modification of \pet{}, enabling us to use it for tasks that require predicting multiple tokens. In extensive experiments, we have identified several factors responsible for the strong performance of \pet{} combined with ALBERT: the possibility to concurrently use multiple patterns for transforming examples into cloze questions, the ability to compensate for patterns that are difficult to understand, the usage of labeled data to perform parameter updates%
% as opposed to using them for priming
, and the underlying LM itself.

We have shown that using \pet{}, it is possible to achieve few-shot text classification
performance similar to \gpt{} on SuperGLUE with LMs that
have three orders of magnitude fewer
parameters. This not only lowers financial cost, but above all reduces environmental impact immensely and leads to a much smaller carbon footprint. We see this as an important contribution to
achieving the goal of
an environmentally more friendly NLP. 
To enable comparisons with our work, we make our code, models and datasets publicly available.

For future work, it would be interesting to see whether \pet{} also works for generative tasks when combined with generative LMs and whether further improvements are possible in multi-task settings.

\paragraph*{Acknowledgments}
This work was funded by the European Research Council (ERC \#740516).
We thank the anonymous reviewers
for their helpful comments.

\bibliography{literatur}

\begin{thebibliography}{60}
\expandafter\ifx\csname natexlab\endcsname\relax\def\natexlab#1{#1}\fi

\bibitem[{Anderson and
  G{\'o}mez-Rodr{\'\i}guez(2020)}]{anderson-gomez-rodriguez-2020-distilling}
Mark Anderson and Carlos G{\'o}mez-Rodr{\'\i}guez. 2020.
\newblock \href {https://doi.org/10.18653/v1/2020.iwpt-1.2} {Distilling neural
  networks for greener and faster dependency parsing}.
\newblock In \emph{Proceedings of the 16th International Conference on Parsing
  Technologies and the IWPT 2020 Shared Task on Parsing into Enhanced Universal
  Dependencies}, pages 2--13, Online. Association for Computational
  Linguistics.

\bibitem[{Beltagy et~al.(2020)Beltagy, Peters, and
  Cohan}]{beltagy2020longformer}
Iz~Beltagy, Matthew~E. Peters, and Arman Cohan. 2020.
\newblock \href {https://arxiv.org/abs/2004.05150} {Longformer: The
  long-document transformer}.
\newblock \emph{Computing Research Repository}, arXiv:2004.05150.

\bibitem[{Brin(1999)}]{brin1999extracting}
Sergey Brin. 1999.
\newblock \href {http://ilpubs.stanford.edu:8090/421/1/1999-65.pdf} {Extracting
  patterns and relations from the world wide web}.
\newblock In \emph{The World Wide Web and Databases}, pages 172--183, Berlin,
  Heidelberg. Springer Berlin Heidelberg.

\bibitem[{Brown et~al.(2020)Brown, Mann, Ryder, Subbiah, Kaplan, Dhariwal,
  Neelakantan, Shyam, Sastry, Askell, Agarwal, Herbert-Voss, Krueger, Henighan,
  Child, Ramesh, Ziegler, Wu, Winter, Hesse, Chen, Sigler, Litwin, Gray, Chess,
  Clark, Berner, McCandlish, Radford, Sutskever, and
  Amodei}]{brown2020language}
Tom Brown, Benjamin Mann, Nick Ryder, Melanie Subbiah, Jared~D Kaplan, Prafulla
  Dhariwal, Arvind Neelakantan, Pranav Shyam, Girish Sastry, Amanda Askell,
  Sandhini Agarwal, Ariel Herbert-Voss, Gretchen Krueger, Tom Henighan, Rewon
  Child, Aditya Ramesh, Daniel Ziegler, Jeffrey Wu, Clemens Winter, Chris
  Hesse, Mark Chen, Eric Sigler, Mateusz Litwin, Scott Gray, Benjamin Chess,
  Jack Clark, Christopher Berner, Sam McCandlish, Alec Radford, Ilya Sutskever,
  and Dario Amodei. 2020.
\newblock \href
  {https://proceedings.neurips.cc/paper/2020/file/1457c0d6bfcb4967418bfb8ac142f64a-Paper.pdf}
  {Language models are few-shot learners}.
\newblock In \emph{Advances in Neural Information Processing Systems},
  volume~33, pages 1877--1901. Curran Associates, Inc.

\bibitem[{Chronopoulou et~al.(2019)Chronopoulou, Baziotis, and
  Potamianos}]{chronopoulou-etal-2019-embarrassingly}
Alexandra Chronopoulou, Christos Baziotis, and Alexandros Potamianos. 2019.
\newblock \href {https://doi.org/10.18653/v1/N19-1213} {An embarrassingly
  simple approach for transfer learning from pretrained language models}.
\newblock In \emph{Proceedings of the 2019 Conference of the North {A}merican
  Chapter of the Association for Computational Linguistics: Human Language
  Technologies, Volume 1 (Long and Short Papers)}, pages 2089--2095,
  Minneapolis, Minnesota. Association for Computational Linguistics.

\bibitem[{Clark et~al.(2019)Clark, Lee, Chang, Kwiatkowski, Collins, and
  Toutanova}]{clark2019boolq}
Christopher Clark, Kenton Lee, Ming-Wei Chang, Tom Kwiatkowski, Michael
  Collins, and Kristina Toutanova. 2019.
\newblock \href {https://doi.org/10.18653/v1/N19-1300} {{B}ool{Q}: Exploring
  the surprising difficulty of natural yes/no questions}.
\newblock In \emph{Proceedings of the 2019 Conference of the North {A}merican
  Chapter of the Association for Computational Linguistics: Human Language
  Technologies, Volume 1 (Long and Short Papers)}, pages 2924--2936,
  Minneapolis, Minnesota. Association for Computational Linguistics.

\bibitem[{Dagan et~al.(2006)Dagan, Glickman, and Magnini}]{dagan2006pascal}
Ido Dagan, Oren Glickman, and Bernardo Magnini. 2006.
\newblock \href
  {https://u.cs.biu.ac.il/~nlp/downloads/publications/RTEChallenge.pdf} {The
  {PASCAL} recognising textual entailment challenge}.
\newblock In \emph{Machine learning challenges. evaluating predictive
  uncertainty, visual object classification, and recognising tectual
  entailment}, pages 177--190. Springer.

\bibitem[{Davison et~al.(2019)Davison, Feldman, and
  Rush}]{davison-etal-2019-commonsense}
Joe Davison, Joshua Feldman, and Alexander Rush. 2019.
\newblock \href {https://doi.org/10.18653/v1/D19-1109} {Commonsense knowledge
  mining from pretrained models}.
\newblock In \emph{Proceedings of the 2019 Conference on Empirical Methods in
  Natural Language Processing and the 9th International Joint Conference on
  Natural Language Processing (EMNLP-IJCNLP)}, pages 1173--1178, Hong Kong,
  China. Association for Computational Linguistics.

\bibitem[{De~Marneffe et~al.(2019)De~Marneffe, Simons, and
  Tonhauser}]{demarneffe:cb}
Marie-Catherine De~Marneffe, Mandy Simons, and Judith Tonhauser. 2019.
\newblock \href
  {https://ojs.ub.uni-konstanz.de/sub/index.php/sub/article/view/601/456} {{The
  CommitmentBank}: Investigating projection in naturally occurring discourse}.
\newblock In \emph{Proceedings of Sinn und Bedeutung 23}.

\bibitem[{Devlin et~al.(2019)Devlin, Chang, Lee, and
  Toutanova}]{devlin2018bert}
Jacob Devlin, Ming-Wei Chang, Kenton Lee, and Kristina Toutanova. 2019.
\newblock \href {https://doi.org/10.18653/v1/N19-1423} {{BERT}: Pre-training of
  deep bidirectional transformers for language understanding}.
\newblock In \emph{Proceedings of the 2019 Conference of the North {A}merican
  Chapter of the Association for Computational Linguistics: Human Language
  Technologies, Volume 1 (Long and Short Papers)}, pages 4171--4186,
  Minneapolis, Minnesota. Association for Computational Linguistics.

\bibitem[{Dodge et~al.(2020)Dodge, Ilharco, Schwartz, Farhadi, Hajishirzi, and
  Smith}]{dodge2020finetuning}
Jesse Dodge, Gabriel Ilharco, Roy Schwartz, Ali Farhadi, Hannaneh Hajishirzi,
  and Noah Smith. 2020.
\newblock \href {https://arxiv.org/abs/2002.06305} {Fine-tuning pretrained
  language models: Weight initializations, data orders, and early stopping}.
\newblock \emph{Computing Research Repository}, arXiv:2002.06305.

\bibitem[{Dogan et~al.(2016)Dogan, Glasmachers, and Igel}]{dogan2016unified}
{\"U}r{\"u}n Dogan, Tobias Glasmachers, and Christian Igel. 2016.
\newblock \href {https://jmlr.org/papers/volume17/11-229/11-229.pdf} {A unified
  view on multi-class support vector classification.}
\newblock \emph{J. Mach. Learn. Res.}, 17(45):1--32.

\bibitem[{Ettinger(2020)}]{ettinger2020bert}
Allyson Ettinger. 2020.
\newblock \href {https://doi.org/10.1162/tacl_a_00298} {What {BERT} is not:
  Lessons from a new suite of psycholinguistic diagnostics for language
  models}.
\newblock \emph{Transactions of the Association for Computational Linguistics},
  8:34–48.

\bibitem[{Ghazvininejad et~al.(2019)Ghazvininejad, Levy, Liu, and
  Zettlemoyer}]{ghazvininejad2019maskpredict}
Marjan Ghazvininejad, Omer Levy, Yinhan Liu, and Luke Zettlemoyer. 2019.
\newblock \href {https://doi.org/10.18653/v1/D19-1633} {Mask-predict: Parallel
  decoding of conditional masked language models}.
\newblock In \emph{Proceedings of the 2019 Conference on Empirical Methods in
  Natural Language Processing and the 9th International Joint Conference on
  Natural Language Processing (EMNLP-IJCNLP)}, pages 6112--6121, Hong Kong,
  China. Association for Computational Linguistics.

\bibitem[{Gong et~al.(2014)Gong, Liu, Yang, and Bourdev}]{gong2014compressing}
Yunchao Gong, Liu Liu, Ming Yang, and Lubomir Bourdev. 2014.
\newblock \href {https://arxiv.org/abs/1412.6115} {Compressing deep
  convolutional networks using vector quantization}.
\newblock \emph{Computing Research Repository}, arXiv:1412.6115.

\bibitem[{Gordon et~al.(2012)Gordon, Kozareva, and
  Roemmele}]{roemmele2011choice}
Andrew Gordon, Zornitsa Kozareva, and Melissa Roemmele. 2012.
\newblock \href {https://www.aclweb.org/anthology/S12-1052} {{S}em{E}val-2012
  task 7: Choice of plausible alternatives: An evaluation of commonsense causal
  reasoning}.
\newblock In \emph{*{SEM} 2012: The First Joint Conference on Lexical and
  Computational Semantics {--} Volume 1: Proceedings of the main conference and
  the shared task, and Volume 2: Proceedings of the Sixth International
  Workshop on Semantic Evaluation ({S}em{E}val 2012)}, pages 394--398,
  Montr{\'e}al, Canada. Association for Computational Linguistics.

\bibitem[{Han et~al.(2016)Han, Mao, and Dally}]{han2015deep_compression}
Song Han, Huizi Mao, and William~J Dally. 2016.
\newblock \href {https://arxiv.org/abs/1510.00149} {Deep compression:
  Compressing deep neural networks with pruning, trained quantization and
  huffman coding}.
\newblock \emph{International Conference on Learning Representations (ICLR)}.

\bibitem[{Han et~al.(2015)Han, Pool, Tran, and Dally}]{NIPS2015_ae0eb3ee}
Song Han, Jeff Pool, John Tran, and William Dally. 2015.
\newblock \href
  {https://proceedings.neurips.cc/paper/2015/file/ae0eb3eed39d2bcef4622b2499a05fe6-Paper.pdf}
  {Learning both weights and connections for efficient neural network}.
\newblock In \emph{Advances in Neural Information Processing Systems},
  volume~28. Curran Associates, Inc.

\bibitem[{Hinton et~al.(2015)Hinton, Vinyals, and Dean}]{hinton2015distilling}
Geoffrey Hinton, Oriol Vinyals, and Jeff Dean. 2015.
\newblock \href {http://arxiv.org/abs/1503.02531} {Distilling the knowledge in
  a neural network}.
\newblock \emph{Computing Research Repository}, arXiv:1503.02531.

\bibitem[{Jiang et~al.(2020)Jiang, Xu, Araki, and Neubig}]{jiang2019know}
Zhengbao Jiang, Frank~F. Xu, Jun Araki, and Graham Neubig. 2020.
\newblock \href {https://doi.org/10.1162/tacl_a_00324} {How can we know what
  language models know?}
\newblock \emph{Transactions of the Association for Computational Linguistics},
  8:423--438.

\bibitem[{Jiao et~al.(2020)Jiao, Yin, Shang, Jiang, Chen, Li, Wang, and
  Liu}]{jiao-etal-2020-tinybert}
Xiaoqi Jiao, Yichun Yin, Lifeng Shang, Xin Jiang, Xiao Chen, Linlin Li, Fang
  Wang, and Qun Liu. 2020.
\newblock \href {https://doi.org/10.18653/v1/2020.findings-emnlp.372}
  {{T}iny{BERT}: Distilling {BERT} for natural language understanding}.
\newblock In \emph{Findings of the Association for Computational Linguistics:
  EMNLP 2020}, pages 4163--4174, Online. Association for Computational
  Linguistics.

\bibitem[{Khashabi et~al.(2018)Khashabi, Chaturvedi, Roth, Upadhyay, and
  Roth}]{khashabi2018looking}
Daniel Khashabi, Snigdha Chaturvedi, Michael Roth, Shyam Upadhyay, and Dan
  Roth. 2018.
\newblock \href {https://doi.org/10.18653/v1/N18-1023} {Looking beyond the
  surface: A challenge set for reading comprehension over multiple sentences}.
\newblock In \emph{Proceedings of the 2018 Conference of the North {A}merican
  Chapter of the Association for Computational Linguistics: Human Language
  Technologies, Volume 1 (Long Papers)}, pages 252--262, New Orleans,
  Louisiana. Association for Computational Linguistics.

\bibitem[{Kitaev et~al.(2020)Kitaev, Kaiser, and Levskaya}]{kitaev2020reformer}
Nikita Kitaev, Lukasz Kaiser, and Anselm Levskaya. 2020.
\newblock \href {https://openreview.net/forum?id=rkgNKkHtvB} {Reformer: The
  efficient transformer}.
\newblock In \emph{International Conference on Learning Representations}.

\bibitem[{Lan et~al.(2020)Lan, Chen, Goodman, Gimpel, Sharma, and
  Soricut}]{lan2019albert}
Zhenzhong Lan, Mingda Chen, Sebastian Goodman, Kevin Gimpel, Piyush Sharma, and
  Radu Soricut. 2020.
\newblock \href {https://openreview.net/forum?id=H1eA7AEtvS} {{ALBERT}: A lite
  {BERT} for self-supervised learning of language representations}.
\newblock In \emph{International Conference on Learning Representations}.

\bibitem[{Levesque et~al.(2011)Levesque, Davis, and
  Morgenstern}]{levesque2011winograd}
Hector~J Levesque, Ernest Davis, and Leora Morgenstern. 2011.
\newblock The {W}inograd schema challenge.
\newblock In \emph{{AAAI} Spring Symposium: Logical Formalizations of
  Commonsense Reasoning}, volume~46, page~47.

\bibitem[{Lewis et~al.(2020)Lewis, Liu, Goyal, Ghazvininejad, Mohamed, Levy,
  Stoyanov, and Zettlemoyer}]{lewis2019bart}
Mike Lewis, Yinhan Liu, Naman Goyal, Marjan Ghazvininejad, Abdelrahman Mohamed,
  Omer Levy, Veselin Stoyanov, and Luke Zettlemoyer. 2020.
\newblock \href {https://doi.org/10.18653/v1/2020.acl-main.703} {{BART}:
  Denoising sequence-to-sequence pre-training for natural language generation,
  translation, and comprehension}.
\newblock In \emph{Proceedings of the 58th Annual Meeting of the Association
  for Computational Linguistics}, pages 7871--7880, Online. Association for
  Computational Linguistics.

\bibitem[{Liu et~al.(2020)Liu, Zhou, Wang, Zhao, Deng, and
  Ju}]{liu-etal-2020-fastbert}
Weijie Liu, Peng Zhou, Zhiruo Wang, Zhe Zhao, Haotang Deng, and Qi~Ju. 2020.
\newblock \href {https://doi.org/10.18653/v1/2020.acl-main.537} {{F}ast{BERT}:
  a self-distilling {BERT} with adaptive inference time}.
\newblock In \emph{Proceedings of the 58th Annual Meeting of the Association
  for Computational Linguistics}, pages 6035--6044, Online. Association for
  Computational Linguistics.

\bibitem[{Liu et~al.(2019)Liu, Ott, Goyal, Du, Joshi, Chen, Levy, Lewis,
  Zettlemoyer, and Stoyanov}]{liu2019roberta}
Yinhan Liu, Myle Ott, Naman Goyal, Jingfei Du, Mandar Joshi, Danqi Chen, Omer
  Levy, Mike Lewis, Luke Zettlemoyer, and Veselin Stoyanov. 2019.
\newblock \href {http://arxiv.org/abs/1907.11692} {{RoBERTa}: {A} robustly
  optimized {BERT} pretraining approach}.
\newblock \emph{Computing Research Repository}, arXiv:1907.11692.

\bibitem[{Mao et~al.(2020)Mao, Wang, Wu, Zhang, Wang, Zhang, Yang, Tong, and
  Bai}]{mao-etal-2020-ladabert}
Yihuan Mao, Yujing Wang, Chufan Wu, Chen Zhang, Yang Wang, Quanlu Zhang, Yaming
  Yang, Yunhai Tong, and Jing Bai. 2020.
\newblock \href {https://doi.org/10.18653/v1/2020.coling-main.287}
  {{L}ada{BERT}: Lightweight adaptation of {BERT} through hybrid model
  compression}.
\newblock In \emph{Proceedings of the 28th International Conference on
  Computational Linguistics}, pages 3225--3234, Barcelona, Spain (Online).
  International Committee on Computational Linguistics.

\bibitem[{McClosky et~al.(2006)McClosky, Charniak, and
  Johnson}]{mcclosky-etal-2006-effective}
David McClosky, Eugene Charniak, and Mark Johnson. 2006.
\newblock \href {https://www.aclweb.org/anthology/N06-1020} {Effective
  self-training for parsing}.
\newblock In \emph{Proceedings of the Human Language Technology Conference of
  the {NAACL}, Main Conference}, pages 152--159, New York City, USA.
  Association for Computational Linguistics.

\bibitem[{Opitz(2019)}]{opitz2019argumentative}
Juri Opitz. 2019.
\newblock \href {https://arxiv.org/abs/1909.09031} {Argumentative relation
  classification as plausibility ranking}.
\newblock In \emph{Preliminary proceedings of the 15th Conference on Natural
  Language Processing (KONVENS 2019): Long Papers}, pages 193--202, Erlangen,
  Germany. German Society for Computational Linguistics \& Language Technology.

\bibitem[{Paszke et~al.(2017)Paszke, Gross, Chintala, Chanan, Yang, DeVito,
  Lin, Desmaison, Antiga, and Lerer}]{paszke2017automatic}
Adam Paszke, Sam Gross, Soumith Chintala, Gregory Chanan, Edward Yang, Zachary
  DeVito, Zeming Lin, Alban Desmaison, Luca Antiga, and Adam Lerer. 2017.
\newblock \href {https://openreview.net/forum?id=BJJsrmfCZ} {Automatic
  differentiation in {PyTorch}}.
\newblock In \emph{NIPS Autodiff Workshop}.

\bibitem[{Petroni et~al.(2019)Petroni, Rocktäschel, Riedel, Lewis, Bakhtin,
  Wu, and Miller}]{Petroni_2019}
Fabio Petroni, Tim Rocktäschel, Sebastian Riedel, Patrick Lewis, Anton
  Bakhtin, Yuxiang Wu, and Alexander Miller. 2019.
\newblock \href {https://doi.org/10.18653/v1/d19-1250} {Language models as
  knowledge bases?}
\newblock \emph{Proceedings of the 2019 Conference on Empirical Methods in
  Natural Language Processing and the 9th International Joint Conference on
  Natural Language Processing (EMNLP-IJCNLP)}.

\bibitem[{Pilehvar and Camacho-Collados(2019)}]{pilehvar2018wic}
Mohammad~Taher Pilehvar and Jose Camacho-Collados. 2019.
\newblock \href {https://doi.org/10.18653/v1/N19-1128} {{W}i{C}: the
  word-in-context dataset for evaluating context-sensitive meaning
  representations}.
\newblock In \emph{Proceedings of the 2019 Conference of the North {A}merican
  Chapter of the Association for Computational Linguistics: Human Language
  Technologies, Volume 1 (Long and Short Papers)}, pages 1267--1273,
  Minneapolis, Minnesota. Association for Computational Linguistics.

\bibitem[{Puri and Catanzaro(2019)}]{puri2019zeroshot}
Raul Puri and Bryan Catanzaro. 2019.
\newblock \href {http://arxiv.org/abs/1912.10165} {Zero-shot text
  classification with generative language models}.
\newblock \emph{Computing Research Repository}, arXiv:1912.10165.

\bibitem[{Radford et~al.(2018)Radford, Narasimhan, Salimans, and
  Sutskever}]{radford2018improving}
Alec Radford, Karthik Narasimhan, Tim Salimans, and Ilya Sutskever. 2018.
\newblock \href
  {https://s3-us-west-2.amazonaws.com/openai-assets/research-covers/language-unsupervised/language_understanding_paper.pdf}
  {Improving language understanding by generative pre-training}.

\bibitem[{Radford et~al.(2019)Radford, Wu, Child, Luan, Amodei, and
  Sutskever}]{radford2018language}
Alec Radford, Jeff Wu, Rewon Child, David Luan, Dario Amodei, and Ilya
  Sutskever. 2019.
\newblock \href
  {https://cdn.openai.com/better-language-models/language_models_are_unsupervised_multitask_learners.pdf}
  {Language models are unsupervised multitask learners}.
\newblock Technical report.

\bibitem[{Raffel et~al.(2020)Raffel, Shazeer, Roberts, Lee, Narang, Matena,
  Zhou, Li, and Liu}]{raffel2019exploring}
Colin Raffel, Noam Shazeer, Adam Roberts, Katherine Lee, Sharan Narang, Michael
  Matena, Yanqi Zhou, Wei Li, and Peter~J. Liu. 2020.
\newblock \href {http://jmlr.org/papers/v21/20-074.html} {Exploring the limits
  of transfer learning with a unified text-to-text transformer}.
\newblock \emph{Journal of Machine Learning Research}, 21(140):1--67.

\bibitem[{Salazar et~al.(2020)Salazar, Liang, Nguyen, and
  Kirchhoff}]{salazar2019masked}
Julian Salazar, Davis Liang, Toan~Q. Nguyen, and Katrin Kirchhoff. 2020.
\newblock \href {https://doi.org/10.18653/v1/2020.acl-main.240} {Masked
  language model scoring}.
\newblock In \emph{Proceedings of the 58th Annual Meeting of the Association
  for Computational Linguistics}, pages 2699--2712, Online. Association for
  Computational Linguistics.

\bibitem[{Sanh et~al.(2019)Sanh, Debut, Chaumond, and
  Wolf}]{sanh2020distilbert}
Victor Sanh, Lysandre Debut, Julien Chaumond, and Thomas Wolf. 2019.
\newblock \href {https://arxiv.org/abs/1910.01108} {{DistilBERT}, a distilled
  version of {BERT}: smaller, faster, cheaper and lighter}.
\newblock In \emph{Proceedings of the 5th Workshop on Energy Efficient Machine
  Learning and Cognitive Computing, NeurIPS 2019}.

\bibitem[{Sanh et~al.(2020)Sanh, Wolf, and Rush}]{NEURIPS2020_eae15aab}
Victor Sanh, Thomas Wolf, and Alexander Rush. 2020.
\newblock \href
  {https://proceedings.neurips.cc/paper/2020/file/eae15aabaa768ae4a5993a8a4f4fa6e4-Paper.pdf}
  {Movement pruning: Adaptive sparsity by fine-tuning}.
\newblock In \emph{Advances in Neural Information Processing Systems},
  volume~33, pages 20378--20389. Curran Associates, Inc.

\bibitem[{Schick and Sch{\"u}tze(2020)}]{schick2019ota}
Timo Schick and Hinrich Sch{\"u}tze. 2020.
\newblock \href {https://arxiv.org/abs/1904.06707} {Rare words: A major problem
  for contextualized embeddings and how to fix it by attentive mimicking}.
\newblock In \emph{Proceedings of the Thirty-Fourth AAAI Conference on
  Artificial Intelligence}.

\bibitem[{Schick and Sch{\"u}tze(2021)}]{schick2020exploiting}
Timo Schick and Hinrich Sch{\"u}tze. 2021.
\newblock \href {https://arxiv.org/abs/2001.07676} {Exploiting cloze questions
  for few shot text classification and natural language inference}.
\newblock In \emph{Proceedings of the 16th Conference of the European Chapter
  of the Association for Computational Linguistics}, Kyiv, Ukraine (Online).
  International Committee on Computational Linguistics.

\bibitem[{Schwartz et~al.(2020{\natexlab{a}})Schwartz, Dodge, Smith, and
  Etzioni}]{schwartz2020green}
Roy Schwartz, Jesse Dodge, Noah~A. Smith, and Oren Etzioni. 2020{\natexlab{a}}.
\newblock \href {https://doi.org/10.1145/3381831} {Green {AI}}.
\newblock \emph{Commun. ACM}, 63(12):54–63.

\bibitem[{Schwartz et~al.(2020{\natexlab{b}})Schwartz, Stanovsky, Swayamdipta,
  Dodge, and Smith}]{schwartz-etal-2020-right}
Roy Schwartz, Gabriel Stanovsky, Swabha Swayamdipta, Jesse Dodge, and Noah~A.
  Smith. 2020{\natexlab{b}}.
\newblock \href {https://doi.org/10.18653/v1/2020.acl-main.593} {The right tool
  for the job: Matching model and instance complexities}.
\newblock In \emph{Proceedings of the 58th Annual Meeting of the Association
  for Computational Linguistics}, pages 6640--6651, Online. Association for
  Computational Linguistics.

\bibitem[{Scudder(1965)}]{scudder1965probability}
H~Scudder. 1965.
\newblock Probability of error of some adaptive pattern-recognition machines.
\newblock \emph{IEEE Transactions on Information Theory}, 11(3):363--371.

\bibitem[{Stock et~al.(2021)Stock, Fan, Graham, Grave, Gribonval, Jegou, and
  Joulin}]{stock2021training}
Pierre Stock, Angela Fan, Benjamin Graham, Edouard Grave, R{\'e}mi Gribonval,
  Herve Jegou, and Armand Joulin. 2021.
\newblock \href {https://openreview.net/forum?id=dV19Yyi1fS3} {Training with
  quantization noise for extreme model compression}.
\newblock In \emph{International Conference on Learning Representations}.

\bibitem[{Strubell et~al.(2019)Strubell, Ganesh, and
  McCallum}]{strubell-etal-2019-energy}
Emma Strubell, Ananya Ganesh, and Andrew McCallum. 2019.
\newblock \href {https://doi.org/10.18653/v1/P19-1355} {Energy and policy
  considerations for deep learning in {NLP}}.
\newblock In \emph{Proceedings of the 57th Annual Meeting of the Association
  for Computational Linguistics}, pages 3645--3650, Florence, Italy.
  Association for Computational Linguistics.

\bibitem[{Talmor et~al.(2020)Talmor, Elazar, Goldberg, and
  Berant}]{talmor2019olmpics}
Alon Talmor, Yanai Elazar, Yoav Goldberg, and Jonathan Berant. 2020.
\newblock \href {https://doi.org/10.1162/tacl_a_00342} {o{LM}pics -- on what
  language model pre-training captures}.
\newblock \emph{Transactions of the Association for Computational Linguistics},
  8:743--758.

\bibitem[{Trinh and Le(2018)}]{trinh2018simple}
Trieu~H. Trinh and Quoc~V. Le. 2018.
\newblock \href {https://arxiv.org/abs/1806.02847} {A simple method for
  commonsense reasoning}.
\newblock \emph{Computing Research Repository}, arXiv:1806.02847.

\bibitem[{Vaswani et~al.(2017)Vaswani, Shazeer, Parmar, Uszkoreit, Jones,
  Gomez, Kaiser, and Polosukhin}]{Vaswani2017}
Ashish Vaswani, Noam Shazeer, Niki Parmar, Jakob Uszkoreit, Llion Jones,
  Aidan~N Gomez, Lukasz Kaiser, and Illia Polosukhin. 2017.
\newblock \href
  {http://papers.nips.cc/paper/7181-attention-is-all-you-need.pdf} {Attention
  is all you need}.
\newblock In \emph{Advances in Neural Information Processing Systems 30}, pages
  5998--6008. Curran Associates, Inc.

\bibitem[{Wang and Cho(2019)}]{wang2019bert}
Alex Wang and Kyunghyun Cho. 2019.
\newblock \href {https://doi.org/10.18653/v1/W19-2304} {{BERT} has a mouth, and
  it must speak: {BERT} as a {M}arkov random field language model}.
\newblock In \emph{Proceedings of the Workshop on Methods for Optimizing and
  Evaluating Neural Language Generation}, pages 30--36, Minneapolis, Minnesota.
  Association for Computational Linguistics.

\bibitem[{Wang et~al.(2019)Wang, Pruksachatkun, Nangia, Singh, Michael, Hill,
  Levy, and Bowman}]{wang2019superglue}
Alex Wang, Yada Pruksachatkun, Nikita Nangia, Amanpreet Singh, Julian Michael,
  Felix Hill, Omer Levy, and Samuel Bowman. 2019.
\newblock \href
  {https://proceedings.neurips.cc/paper/2019/file/4496bf24afe7fab6f046bf4923da8de6-Paper.pdf}
  {{SuperGLUE}: A stickier benchmark for general-purpose language understanding
  systems}.
\newblock In \emph{Advances in Neural Information Processing Systems},
  volume~32. Curran Associates, Inc.

\bibitem[{Weston and Watkins(1999)}]{weston1999support}
Jason Weston and Chris Watkins. 1999.
\newblock \href
  {http://dblp.uni-trier.de/db/conf/esann/esann1999.html#WestonW99} {Support
  vector machines for multi-class pattern recognition}.
\newblock In \emph{ESANN}, volume~99, pages 219--224.

\bibitem[{Wolf et~al.(2020)Wolf, Debut, Sanh, Chaumond, Delangue, Moi, Cistac,
  Rault, Louf, Funtowicz, Davison, Shleifer, von Platen, Ma, Jernite, Plu, Xu,
  Le~Scao, Gugger, Drame, Lhoest, and Rush}]{wolf2019transformers}
Thomas Wolf, Lysandre Debut, Victor Sanh, Julien Chaumond, Clement Delangue,
  Anthony Moi, Pierric Cistac, Tim Rault, Remi Louf, Morgan Funtowicz, Joe
  Davison, Sam Shleifer, Patrick von Platen, Clara Ma, Yacine Jernite, Julien
  Plu, Canwen Xu, Teven Le~Scao, Sylvain Gugger, Mariama Drame, Quentin Lhoest,
  and Alexander Rush. 2020.
\newblock \href {https://www.aclweb.org/anthology/2020.emnlp-demos.6}
  {Transformers: State-of-the-art natural language processing}.
\newblock In \emph{Proceedings of the 2020 Conference on Empirical Methods in
  Natural Language Processing: System Demonstrations}, pages 38--45, Online.
  Association for Computational Linguistics.

\bibitem[{Xin et~al.(2020)Xin, Nogueira, Yu, and Lin}]{xin-etal-2020-early}
Ji~Xin, Rodrigo Nogueira, Yaoliang Yu, and Jimmy Lin. 2020.
\newblock \href {https://doi.org/10.18653/v1/2020.sustainlp-1.11} {Early
  exiting {BERT} for efficient document ranking}.
\newblock In \emph{Proceedings of SustaiNLP: Workshop on Simple and Efficient
  Natural Language Processing}, pages 83--88, Online. Association for
  Computational Linguistics.

\bibitem[{Yang et~al.(2019)Yang, Dai, Yang, Carbonell, Salakhutdinov, and
  Le}]{NIPS2019_8812}
Zhilin Yang, Zihang Dai, Yiming Yang, Jaime Carbonell, Russ~R Salakhutdinov,
  and Quoc~V Le. 2019.
\newblock \href
  {http://papers.nips.cc/paper/8812-xlnet-generalized-autoregressive-pretraining-for-language-understanding.pdf}
  {Xlnet: Generalized autoregressive pretraining for language understanding}.
\newblock In H.~Wallach, H.~Larochelle, A.~Beygelzimer, F.~d\textquotesingle
  Alch\'{e}-Buc, E.~Fox, and R.~Garnett, editors, \emph{Advances in Neural
  Information Processing Systems 32}, pages 5753--5763. Curran Associates, Inc.

\bibitem[{Yarowsky(1995)}]{yarowsky-1995-unsupervised}
David Yarowsky. 1995.
\newblock \href {https://doi.org/10.3115/981658.981684} {Unsupervised word
  sense disambiguation rivaling supervised methods}.
\newblock In \emph{33rd Annual Meeting of the Association for Computational
  Linguistics}, pages 189--196, Cambridge, Massachusetts, USA. Association for
  Computational Linguistics.

\bibitem[{Zafrir et~al.(2019)Zafrir, Boudoukh, Izsak, and
  Wasserblat}]{Zafrir2019Q8BERTQ8}
Ofir Zafrir, Guy Boudoukh, Peter Izsak, and Moshe Wasserblat. 2019.
\newblock \href {https://arxiv.org/abs/1910.06188} {{Q8BERT}: Quantized 8bit
  {BERT}}.
\newblock In \emph{NeurIPS EMC2 Workshop}.

\bibitem[{Zhang et~al.(2018)Zhang, Liu, Liu, Gao, Duh, and
  Durme}]{zhang2018record}
Sheng Zhang, Xiaodong Liu, Jingjing Liu, Jianfeng Gao, Kevin Duh, and
  Benjamin~Van Durme. 2018.
\newblock \href {https://arxiv.org/abs/1810.12885} {{ReCoRD}: Bridging the gap
  between human and machine commonsense reading comprehension}.
\newblock \emph{Computing Research Repository}, arXiv:1810.12885.

\end{thebibliography}
\bibliographystyle{acl_natbib}

\appendix

\section{Training Details}
\label{appendix:training_details}

Our implementation can be found in the supplementary material. It extends the original implementation of \pet{} by \citet{schick2020exploiting} which, in turn, is based on the Transformers library \citep{wolf2019transformers} and PyTorch \citep{paszke2017automatic}. All dependencies are listed in \texttt{requirements.txt}. Detailed instructions on how our results can be reproduced using this implementation can be found in \texttt{README.md}.

Unless explicitly stated differently, we use the exact same set of hyperparameters as \citet{schick2020exploiting} (Table~\ref{hyperparameters-table}) with the only difference that for \ipet{}, we only train 3 generations of models to speed up training. 
All of our experiments were conducted using a single GPU with 11GB RAM
(NVIDIA GeForce GTX 1080 Ti). With this GPU, training a single \pet{} model for 250 steps took approximately 45 minutes. Depending on the task, labeling
unlabeled examples took 0.2--1.5 hours per model.
Training the final classifier for 5,000 steps on the
soft-labeled dataset took 2.5 hours on average.
Below, we list task-specific implementation details for all tasks in SuperGLUE.

\paragraph{COPA} For COPA, we randomly switch the two options $c_1$ and $c_2$ during training with a probability of 50\% to make the input more diverse; for inference, we always keep the original order. For  distilling the final \pet{} model, we obtain logits for unlabeled examples $x$ from individual PVPs $\mathbf{p}$ as $s_\mathbf{p}(y \mid x ) = \log q_\mathbf{p}(y \mid x)$; we use the input format proposed by \citet{liu2019roberta}.

\paragraph{WiC} Similar to COPA, we randomly switch the input sentences $s_1$ and $s_2$ during training. Given a word $w$ and two sentences $s_1$ and $s_2$, we use the sequence $w$: $s_1\ |\ s_2$ as input for the final sequence classification model, where $|$ marks the boundary between two text segments.

\paragraph{WSC} Unlike other SuperGLUE tasks, the WSC formulation of \citet{raffel2019exploring} and \citet{brown2020language} requires free-form completion, meaning that for each sentence $s$ and pronoun $p$, we only have a single correct choice $n$ that the model needs to predict, but we do not provide any alternatives. During training, we thus use regular cross entropy loss between $n$ and $\tilde{q}_\mathbf{p}(n \mid s, p)$ as defined in Eq.~4. However, in many cases this would allow the LM to easily identify the correct target based on the number of masks provided, so we modify each target by randomly adding up to three additional mask tokens, for which we require the model to predict a special \texttt{<pad>} token. For inference, we always just add a single mask token to ensure consistent results across multiple evaluations and perform greedy decoding as described in Section~3. We then follow \citet{raffel2019exploring} to map the output produced by the LM to a label $y \in \{ \text{true}, \text{false} \}$. For distillation, given an unlabeled example $x$ we set $s_\mathbf{p}(y \mid x ) = 1$ if the model's output for $x$ was mapped to $y$ and $s_\mathbf{p}(y \mid x) = 0$ otherwise. We provide inputs to the final \pet{} model in the format $s\ |\ n$ where $|$ is the boundary between two text segments and mark $p$ in $s$ with asterisks.

\paragraph{MultiRC} Deviating from the hyperparameters used by \citet{schick2020exploiting}, we use a maximum sequence length of 512 tokens for MultiRC both during training and inference because we found many passages to be much longer than 256 tokens. Input for the final sequence classification model is of the form $p\ |\ q\ |\ a$ where $p$ is the passage, $q$ is the question, $a$ is the answer candidate and we use $|$ to mark boundaries between text segments.

\paragraph{ReCoRD} For ReCoRD, we again use a maximum sequence length of 512 because many passages require more than 256 tokens. For some questions $q$, the ReCoRD training set contains a huge number of answer candidates. To facilitate training, we split each example into multiple examples as follows: let $C$ be the set of answer candidates with $C^+ \subset C$ being the set of correct answers. We create a training example for each $c \in C^+$ by randomly selecting up to 9 negative examples from $C \setminus C^+$ for a total of 10 answer candidates.

\begin{table}
	\small
	\centering
	\begin{tabularx}{\linewidth}{Xl}
		\toprule
		\textbf{Parameter} & \textbf{Value} \\
		\midrule
		\texttt{adam\_epsilon} 					& 1e-8  \\
		\texttt{gradient\_accumulation\_steps} 	& 8     \\
		\texttt{learning\_rate} 				& 1e-5  \\
		\texttt{max\_grad\_norm} 				& 1.0   \\
		\texttt{max\_seq\_length} 				& 256   \\
		\texttt{pet\_max\_steps}  				& 250 \\
		\texttt{sc\_max\_steps} 				& 5,000 \\
		\texttt{per\_gpu\_train\_batch\_size} 	& 2 \\
		\texttt{distillation\_temperature} 		& 2  \\
		\texttt{weight\_decay} 					& 0.01 \\
		\bottomrule
	\end{tabularx}
	\caption{Hyperparameters for \textsc{Pet} from \citet{schick2020exploiting}}
	\label{hyperparameters-table}
\end{table}

\section{Dataset Details}

\begin{table}
	\small
	\centering
	\begin{tabularx}{\linewidth}{lXrrr}
		\toprule
		\textbf{Dataset} &\textbf{Metrics} & \textbf{$|$Unlabeled$|$} & \textbf{$|$Dev$|$} & \textbf{$|$Test$|$} \\
		\midrule
		BoolQ & Acc.  & 9,427 & 3,270 & 3,245 \\
		CB & Acc./F1 & 20,000 & 57 & 250 \\
		COPA & Acc. & 400 & 100 & 500 \\
		MultiRC & F1$_\text{a}$/EM & 5,100 & 953 & 1,800 \\
		ReCoRD & F1/EM & 20,000 & 10,000 & 10,000 \\
		RTE & Acc. & 20,000 & 278 & 300 \\
		WiC & Acc. & 6,000 & 638 & 1,400 \\
		WSC & Acc. & 554 & 104 & 146 \\
		\bottomrule
	\end{tabularx}
	\caption{Important statistics for all datasets used}
	\label{datasets-table}
\end{table}

For each task and number of examples $t$, we create the FewGLUE training set $\mathcal{T}$ by shuffling the entire original training set with a fixed random seed and collecting the first $32$ examples of the shuffled dataset.
Following  \citep{raffel2019exploring,brown2020language}, we select only positive examples for WSC; for both MultiRC and ReCoRD, we follow \citet{brown2020language} and select a total of 32 questions -- which corresponds to more than 32 training examples -- to enable a fair comparison with \gpt{}.

The unlabeled datasets for all tasks are obtained by collecting up to $20,000$ examples from their training sets and removing the labels. As the training sets for RTE and CB are very small, for both tasks we additionally select random unlabeled examples from the MNLI training set for a total of $20,000$ examples. For evaluation, we use the official validation and test sets for all tasks that are available at \url{https://super.gluebenchmark.com/tasks}. All datasets included in SuperGLUE are in English. Additional details for each dataset are given in Table~\ref{datasets-table}.

\paragraph{Preprocessing} We do not perform any preprocessing, except shortening all examples to the maximum sequence length. This is done using the \emph{longest first} strategy implemented in the Transformers library. All input sequences are truncated \emph{before} applying patterns.

\end{document}